\documentclass[10pt,journal,compsoc]{IEEEtran}

\usepackage{color,xcolor}
\usepackage{epsfig}
\usepackage{graphicx}
\usepackage{subfiles}

\usepackage{adjustbox}
\usepackage{array}
\usepackage{booktabs}
\usepackage{colortbl}
\usepackage{float,wrapfig}
\usepackage{hhline}
\usepackage{multirow}
\usepackage{subcaption} 
\usepackage[labelfont={bf},labelsep={period},font={small}]{caption}

\let\llncssubparagraph\subparagraph
\let\subparagraph\paragraph
\usepackage[compact]{titlesec}
\let\subparagraph\llncssubparagraph

\usepackage{amsmath,amsfonts,amssymb}
\usepackage{bm}
\usepackage{nicefrac}
\usepackage{microtype}

\usepackage{changepage}
\usepackage{extramarks}
\usepackage{fancyhdr}
\usepackage{lastpage}
\usepackage{setspace}
\usepackage{soul}
\usepackage{xspace}

\usepackage{url}

\usepackage{algorithm}
\usepackage{algpseudocode}
\usepackage{enumerate}
\usepackage{times}

\usepackage[pagebackref=true,breaklinks=true,letterpaper=true,bookmarks=false]{hyperref}
\usepackage{pbox}
\usepackage{footnote}
\usepackage{tablefootnote}

\usepackage{paralist}

\usepackage{enumitem}
\setitemize{noitemsep,topsep=0pt,parsep=0pt,partopsep=0pt}
\usepackage{paralist}
\newcommand{\netFull}{TSM\xspace}
\newcommand{\netHead}{TSM}

\newcommand{\method}{Temporal Shift\xspace}

\makeatletter
\newcommand\footnoteref[1]{\protected@xdef\@thefnmark{\ref{#1}}\@footnotemark}
\makeatother

\newcommand{\itddd}{I3D\textsubscript{3$\times$3$\times$3}\xspace}
\newcommand{\itd}{I3D\textsubscript{3$\times$1$\times$1}\xspace}
\newcommand{\tsm}{TSM\textsubscript{8f}\xspace}

\makeatletter
\DeclareRobustCommand\onedot{\futurelet\@let@token\@onedot}
\def\@onedot{\ifx\@let@token.\else.\null\fi\xspace}

\def\eg{\emph{e.g}\onedot} 
\def\ie{\emph{i.e}\onedot} 
 
\def\etc{\emph{etc}\onedot} \def\vs{\emph{vs}\onedot}
 
\def\etal{\emph{et al}\onedot}

\makeatother

\ifCLASSOPTIONcompsoc
  \usepackage[nocompress]{cite}
\else
  \usepackage{cite}
\fi

\ifCLASSINFOpdf
\else
\fi

\begin{document}

\title{TSM: Temporal Shift Module for Efficient and Scalable Video Understanding on Edge Devices}

\author{Ji~Lin, Chuang~Gan, Kuan~Wang
        and~Song~Han
\IEEEcompsocitemizethanks{\IEEEcompsocthanksitem J. Lin, K. Wang, S. Han are with the Department
of Electrical Engineering and Computer Science, Massachusetts Institute of Technology.\protect\\
E-mail: \{jilin, kuanwang, songhan\}@mit.edu
\IEEEcompsocthanksitem C. Gan is with MIT-IBM Watson AI Lab.\protect\\
E-mail: ganchuang@csail.mit.edu
}
}

\markboth{Journal of \LaTeX\ Class Files,~Vol.~14, No.~8, August~2015}%
{Lin \MakeLowercase{\textit{et al.}}: TSM: Temporal Shift Module for Efficient Video Understanding}

\IEEEtitleabstractindextext{%
\begin{abstract}

The explosive growth in video streaming requires video understanding at high accuracy and low computation cost. Conventional 2D CNNs are computationally cheap but cannot capture temporal relationships; 3D CNN based methods can achieve good performance but are computationally intensive. In this paper, we propose a generic and effective Temporal Shift Module (TSM) that enjoys both high efficiency and high performance. 
The key idea of TSM is to shift part of the channels along the temporal dimension, thus facilitate information exchanged among neighboring frames. It can be inserted into 2D CNNs to achieve temporal modeling at zero computation and zero parameters. TSM offers several unique advantages. Firstly, TSM has high performance; it ranks the first on the Something-Something leaderboard upon submission. Secondly, TSM has high efficiency; it achieves a high frame rate of 74fps and 29fps for online video recognition on Jetson Nano and Galaxy Note8. Thirdly, TSM has higher scalability compared to 3D networks, enabling large-scale Kinetics training on 1,536 GPUs in 15 minutes. Lastly, TSM enables action concepts learning, which 2D networks cannot model; we visualize the category attention map and find that spatial-temporal action detector emerges during the training of classification tasks. The code is publicly available at \url{https://github.com/mit-han-lab/temporal-shift-module}.


\end{abstract}

}

\maketitle

\IEEEdisplaynontitleabstractindextext

%
\IEEEpeerreviewmaketitle

\IEEEraisesectionheading{\section{Introduction}\label{sec:introduction}}
\IEEEPARstart{H}{ardware-efficient} video understanding is an important step towards real-world deployment, both on the cloud and on the edge. For example, there are over $10^5$ hours of videos uploaded to YouTube every day to be processed for recommendation and ads ranking; tera-bytes of sensitive videos in hospitals need to be processed locally on edge devices to protect privacy. 
All these industry applications require both accurate and efficient video understanding.




Deep learning has become the standard for video understanding over the years~\cite{tran2015learning, wang2016temporal, carreira2017quo, wang2017non, zolfaghari2018eco, xie2018rethinking, zhou2017temporal}. One key difference between video recognition and image recognition is the need for \emph{temporal modeling}. For example, to distinguish between opening and closing a box, reversing the order will give opposite results, so temporal modeling is critical. Existing efficient video understanding approaches directly use 2D CNN~\cite{karpathy2014large, simonyan2014two, wang2016temporal, zhou2017temporal}. However, 2D CNN on individual frames could not capture the temporal information very well.
3D CNNs~\cite{tran2015learning, carreira2017quo} can jointly learn spatial and temporal features but the computation cost is large, making the deployment on edge devices difficult; it cannot be applied to real-time online video recognition. 
There are works to trade off between temporal modeling and computation, such as post-hoc fusion~\cite{girdhar2017actionvlad, feichtenhofer2016convolutional, zhou2017temporal, donahue2015long} and mid-level temporal fusion~\cite{zolfaghari2018eco, xie2018rethinking, tran2018closer}. Such methods sacrifice the low-level temporal modeling for efficiency, but much of the useful information is lost during the feature extraction before the temporal fusion happens.
 



\begin{figure}[h]
\centering
\begin{subfigure}[b]{0.149\textwidth}
	\includegraphics[width=\textwidth]{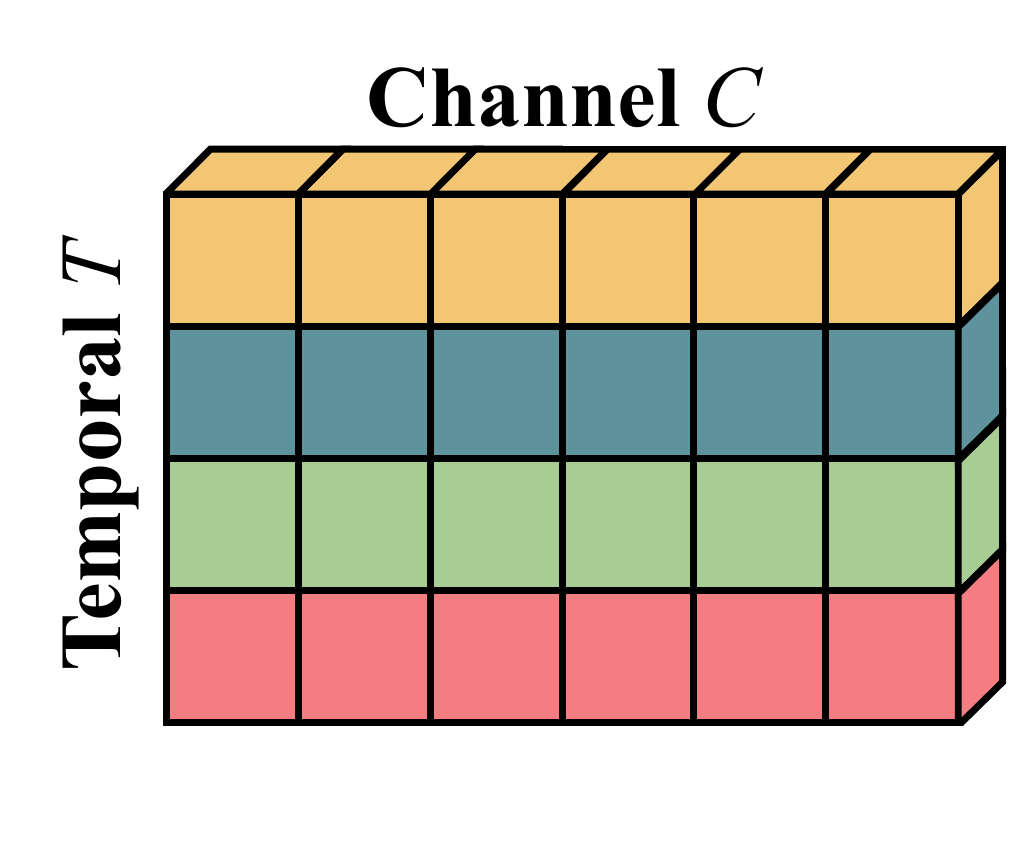}
	\caption{The original tensor without shift.
}
	\label{fig:shift_ori}
\end{subfigure}
~
\begin{subfigure}[b]{0.149\textwidth}
	\includegraphics[width=\textwidth]{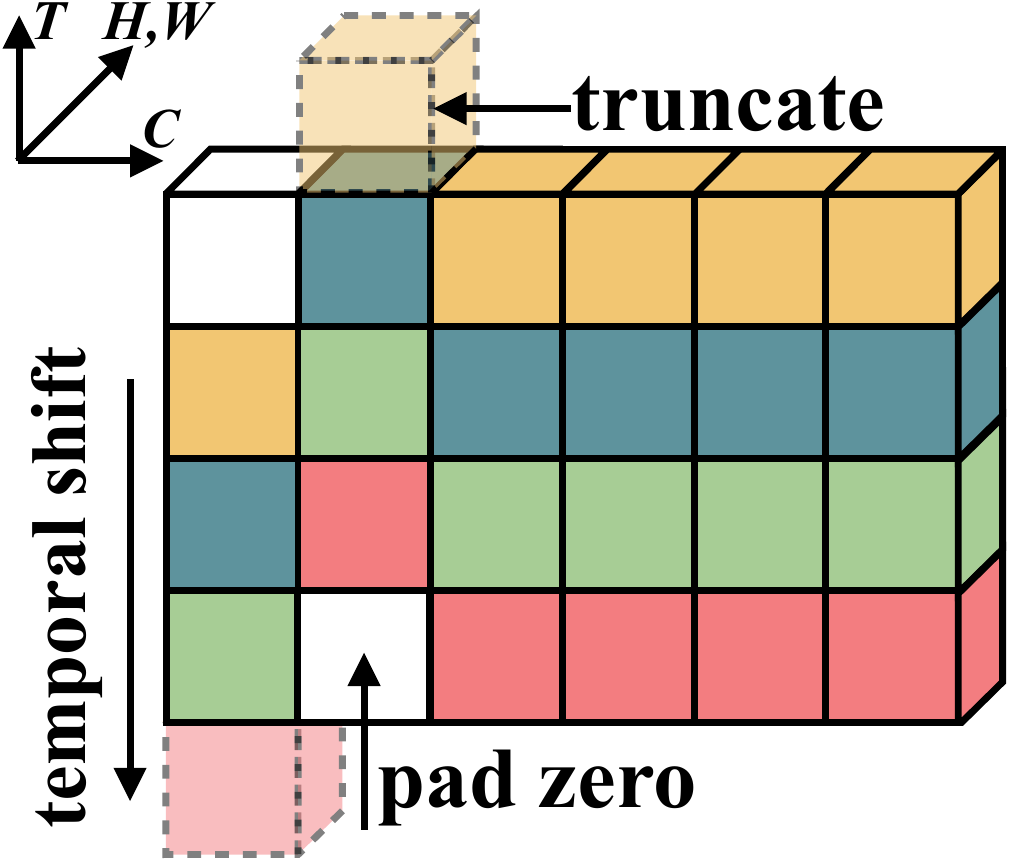}
	\caption{Offline temporal shift (bi-direction). 
}
	\label{fig:shift_zero}
\end{subfigure}
~
\begin{subfigure}[b]{0.149\textwidth}
	\includegraphics[width=\textwidth]{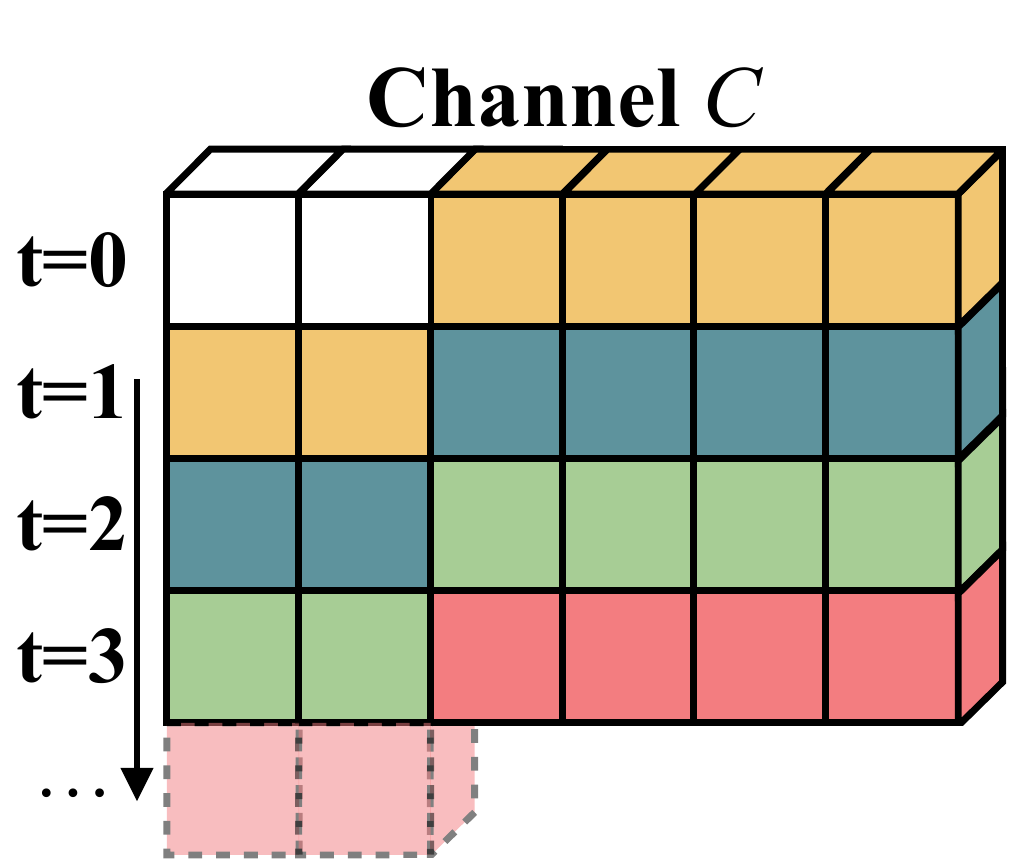}
	\caption{Online temporal shift (uni-direction). 
}
	\label{fig:shift_online}
\end{subfigure}
\caption{\textbf{Temporal Shift Module (TSM)} performs efficient temporal modeling by moving the feature map along the temporal dimension. It is computationally free on top of a 2D convolution, but achieves strong temporal modeling ability.
TSM efficiently supports both \textbf{offline} and \textbf{online} video recognition.
Bi-directional TSM mingles both past and future frames with the current frame, which is suitable for high-throughput offline video recognition. Uni-directional TSM mingles only the past frame with the current frame, which is suitable for low-latency online video recognition. 
}
\label{fig:shift}
\end{figure}
In this paper, we propose a new perspective for efficient temporal modeling in video understanding by proposing a novel Temporal Shift Module (TSM). Concretely, an activation in a video model can be represented as $A\in \mathbb{R}^{N\times C\times T \times H \times W}$,  where $N$ is the batch size, $C$ is the number of channels, $T$ is the temporal dimension, $H$ and $W$ are the spatial resolutions. Traditional 2D CNNs operate independently over the dimension $T$; thus no temporal modeling takes effects (Figure~\ref{fig:shift_ori}).  In contrast, our Temporal Shift Module (TSM) shifts the channels along the temporal dimension, both forward and backward. As shown in Figure~\ref{fig:shift_zero}, the information from neighboring frames is mingled with the current frame after shifting. Our intuition is: the convolution operation consists of \emph{shift} and \emph{multiply-accumulate}. We \emph{shift} in the time dimension by $\pm1$ and fold the \emph{multiply-accumulate} from time dimension to channel dimension. For real-time online video understanding, future frames can't get shifted to the present, so we use a uni-directional TSM (Figure~\ref{fig:shift_online}) to perform online video understanding. 

Despite the zero-computation nature of the shift operation, we empirically find that simply adopting the spatial shift strategy~\cite{wu2017shift} used in image classifications introduces two major issues for video understanding:
(1) it is \emph{not efficient}: shift operation is conceptually zero FLOP but incurs data movement. The additional cost of data movement is non-negligible and will result in latency increase. This phenomenon has been exacerbated in the video networks since they usually have a large memory consumption (5D activation). 
(2) It is \emph{not accurate}: shifting too many channels in a network will significantly hurt the spatial modeling ability and result in performance degradation.  To tackle the problems, we make two technical contributions. (1) We use a \emph{temporal partial shift} strategy: instead of shifting all the channels, we shift only a small portion of the channels for efficient temporal fusion. Such strategy significantly cuts down the data movement cost (Figure~\ref{fig:tsm_discuss_a}).
 (2) We insert TSM inside \emph{residual branch} rather than outside so that the activation of the current frame is preserved, which does not harm the spatial feature learning capability of the 2D CNN backbone.

To verify the effectiveness of TSM, we carried out comprehensive experiments: (1) we show that TSM consistently improve the video recognition performance compared to 2D model without incurring extra computation or parameters; it also achieves state-of-the-art performance on multiple action recognition dataset; (2) TSM has much better accuracy-computation trade-off compared to prior works.
The contributions of our paper are summarized as follows:





\begin{itemize}
\item We provide a new perspective for efficient video model design by temporal shift, which is computationally free but has strong spatio-temporal modeling ability. 

\item We observed that naive shift cannot achieve high efficiency or high performance.
We then proposed two technical modifications \emph{partial shift} and \emph{residual shift} to realize a high efficiency model design.

\item We propose  \emph{bi-directional TSM} for \emph{offline} video understanding that achieves state-of-the-art performance. It ranks the first on Something-Something leaderboard upon publication.


\item We propose \emph{uni-directional TSM} for \emph{online} real-time video recognition with strong temporal modeling capacity at low latency on edge devices.

\item With the efficient design of TSM, we scale up the training of video network to 1,536 GPUs, and finish the training on Kinetics dataset in 15 minutes, without losing accuracy. To the best of our knowledge, we are the first to systematically investigate the large-scale training on video recognition.

\item We provide an in-depth analysis to understand the learned knowledge inside the TSM module, and find that spatial-temporal action detector automatically emerges during training using only classification labels. 

\end{itemize}

\section{Related Work}\label{sec:relatedwork}
In this section, we briefly review four related topics: 1) Deep Video Recognition, 2) Temporal Modeling, and 3)  Efficient Neural Networks.
\subsection{Deep Video Recognition}

\subsubsection{2D CNN.}
Using the 2D CNN is a straightforward way to conduct video recognition~\cite{karpathy2014large, simonyan2014two, wang2016temporal, gan2015devnet, feichtenhofer2016spatiotemporal, feichtenhofer2016convolutional, bilen2016dynamic}. For example, Simonyan~\etal~\cite{simonyan2014two} designed a two-stream CNN for RGB input (spatial stream) and optical flow~\cite{zach2007duality} input (temporal stream) respectively. Temporal Segment Networks (TSN)~\cite{wang2016temporal} extracted averaged features from strided sampled frames. Such methods are more efficient compared to 3D counterparts but cannot infer the temporal order or more complicated temporal relationships. 

\subsubsection{3D CNN.}
3D convolutional neural networks can jointly learn spatio-temporal features. Tran~\etal~\cite{tran2015learning} proposed a 3D CNN based on VGG models, named C3D, to learn spatio-temporal features from a frame sequence. Carreira and Zisserman~\cite{carreira2017quo} proposed to inflate all the 2D convolution filters in an Inception V1 model~\cite{szegedy2015going} into 3D convolutions. However, 3D CNNs are computationally heavy, making the deployment difficult. They also have more parameters than 2D counterparts, thus are more prone to over-fitting. On the other hand, our \netFull has the same spatial-temporal modeling ability as 3D CNN while enjoying the same computation and parameters as the 2D CNNs.

\subsubsection{Trade-offs.}
There have been attempts to trade off expressiveness and computation costs. Lee~\etal~\cite{lee2018motion} proposed a motion filter to generate spatio-temporal features from 2D CNN.
Tran~\etal~\cite{tran2018closer} and Xie~\etal~\cite{xie2018rethinking} proposed to study mixed 2D and 3D networks, either first using 3D and later 2D (bottom-heavy) or first 2D and later 3D (top-heavy) architecture. ECO~\cite{zolfaghari2018eco} also uses a similar top-heavy architecture to achieve a very efficient framework. Another way to save computation is to decompose the 3D convolution into a 2D spatial convolution and a 1D temporal convolution~\cite{tran2018closer, qiu2017learning, sun2015human}. For mixed 2D-3D CNNs, they still need to remove low-level temporal modeling or high-level temporal modeling. Compared to decomposed convolutions, our method completely removes the computation cost of temporal modeling has enjoys better hardware efficiency.

\subsection{Temporal Modeling}
A direct way for temporal modeling is to use 3D CNN based methods as discussed above. Wang~\etal~\cite{wang2017non} proposed a spatial-temporal non-local module to capture long-range dependencies. Wang~\etal~\cite{wang2018videos} proposed to represent videos as space-time region graphs. 
An alternative way to model the temporal relationships is to use 2D CNN + post-hoc fusion~\cite{girdhar2017actionvlad, feichtenhofer2016convolutional, zhou2017temporal, donahue2015long}. Some works use LSTM~\cite{hochreiter1997long} to aggregate the 2D CNN features~\cite{yue2015beyond, donahue2015long, srivastava2015unsupervised, gan2016webly, gan2016you}.
Attention mechanism also proves to be effective for temporal modeling~\cite{sharma2015action, li2018videolstm, long2018attention}.
Zhou~\etal~\cite{zhou2017temporal} proposed Temporal Relation Network to learn and reason about temporal dependencies.
The former category is computational heavy, while the latter cannot capture the useful low-level information that is lost during feature extraction. Our method offers an efficient solution at the cost of 2D CNNs, while enabling both low-level and high-level temporal modeling, just like 3D-CNN based methods.

\subsection{Efficient Neural Networks}
The efficiency of 2D CNN has been extensively studied. Some works focused on designing an efficient model~\cite{iandola2016squeezenet, howard2017mobilenets, sandler2018mobilenetv2, zhang2017shufflenet}. Recently neural architecture search~\cite{zoph2016neural, zoph2017learning, liu2017progressive} has been introduced to find an efficient architecture automatically~\cite{tan2018mnasnet, cai2018proxylessnas}. Another way is to prune, quantize and compress an existing model for efficient deployment~\cite{han2015learning, han2015deep, lin2017runtime, zhu2016trained, he2018amc, wang2018haq}. Address shift, which is a hardware-friendly primitive, has also been exploited for compact 2D CNN design on image recognition tasks~\cite{wu2017shift, zhong2018shift}. Nevertheless, 
we observe that directly adopting the shift operation on video recognition task neither maintains efficiency nor accuracy, due to the complexity of the video data.

\section{\method Module (TSM)}\label{tsm_module}
We first explain the intuition behind TSM: data movement and computation can be separated in a convolution. However, we observe that such naive shift operation neither achieves high efficiency nor high performance. To tackle the problem, we propose two techniques minimizing the data movement and increasing the model capacity, which leads to the efficient TSM module.

\subsection{Intuition}

Let us first consider a normal convolution operation. For brevity, we used a 1-D convolution with the kernel size of 3 as an example. Suppose the weight of the convolution is $W=(w_1, w_2, w_3)$, and the input $X$ is a 1-D vector with infinite length. The convolution operator $Y=\text{Conv}(W, X)$ can be written as:
$ Y_i =  w_1 X_{i-1} + w_2 X_{i} + w_3 X_{i+1}$.
We can decouple the operation of convolution into two steps: \emph{shift} and \emph{multiply-accumulate}: we shift the input $X$ by $-1, 0, +1$ and multiply by $w_1, w_2, w_3$ respectively, which sum up to be $Y$. Formally, the \emph{shift} operation is:
\begin{equation}
\vspace{-5pt}
    X^{-1}_i = X_{i-1},~~~~ X^{0}_i = X_i, ~~~~ X^{+1}_i = X_{i+1}
\end{equation}
and the \emph{multiply-accumulate} operation is:
\begin{equation}
\vspace{-5pt}
    Y=w_1 X^{-1} + w_2 X^{0} + w_3 X^{+1}
\end{equation}
The first step \emph{shift} can be conducted without any multiplication. While the second step is more computationally expensive, our \method module \emph{merges} the \emph{multiply-accumulate} into the following 2D convolution, so 
it introduces no extra cost compared to 2D CNN based models.

The proposed \method module is described in Figure~\ref{fig:shift}.
In Figure~\ref{fig:shift_ori}, we describe a tensor with $C$ channels and $T$ frames. The features at different time stamps are denoted as different colors in each row. 
Along the temporal dimension, we shift part of the channels by $-1$, another part by $+1$, leaving the rest un-shifted (Figure~\ref{fig:shift_zero}).
For online video recognition setting, we also provide an online version of TSM (Figure~\ref{fig:shift_online}). In the online setting, we cannot access future frames, therefore, we only shift from past frames to future frames in a uni-directional fashion.

\subsection{Naive Shift Does Not Work}\label{sec:place_proportion}

Despite the simple philosophy behind the proposed module, we find that directly applying the spatial shift strategy~\cite{wu2017shift} to the temporal dimension cannot provide high performance nor efficiency. To be specific, if we shift all or most of the channels, it brings two disasters:  \textbf{(1) Worse efficiency due to large data movement}. The shift operation enjoys no computation, but it involves data movement. 
Data movement increases the memory footprint and inference latency on hardware. Worse still, such effect is exacerbated in the video understanding networks due to large activation size (5D tensor). When using the naive shift strategy shifting every map, we observe a 13.7\% increase in CPU latency and 12.4\% increase in GPU latency, making the overall inference slow. \textbf{(2) Performance degradation due to worse spatial modeling ability.} By shifting part of the channels to neighboring frames, the information contained in the channels is no longer accessible for the current frame, which may harm the spatial modeling ability of the 2D CNN backbone. We observe a 2.6\% accuracy drop when using the naive shift implementation compared to the 2D CNN baseline (TSN).

\subsection{ Module Design}

\begin{figure}[t]
\centering
\begin{subfigure}[b]{0.23\textwidth}
    	\includegraphics[width=\textwidth]{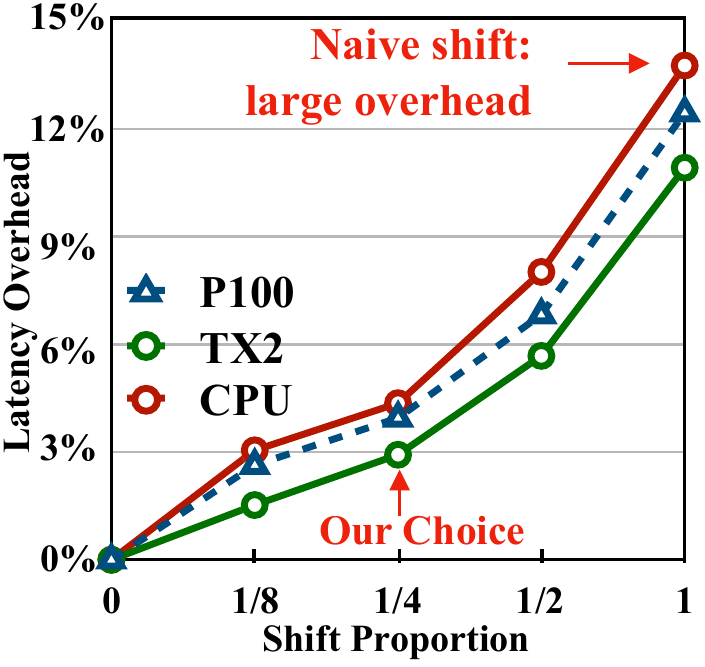}
	\caption{Overhead \vs proportion.}
	\label{fig:tsm_discuss_a}
\end{subfigure}
\begin{subfigure}[b]{0.23\textwidth}
	\includegraphics[width=\textwidth]{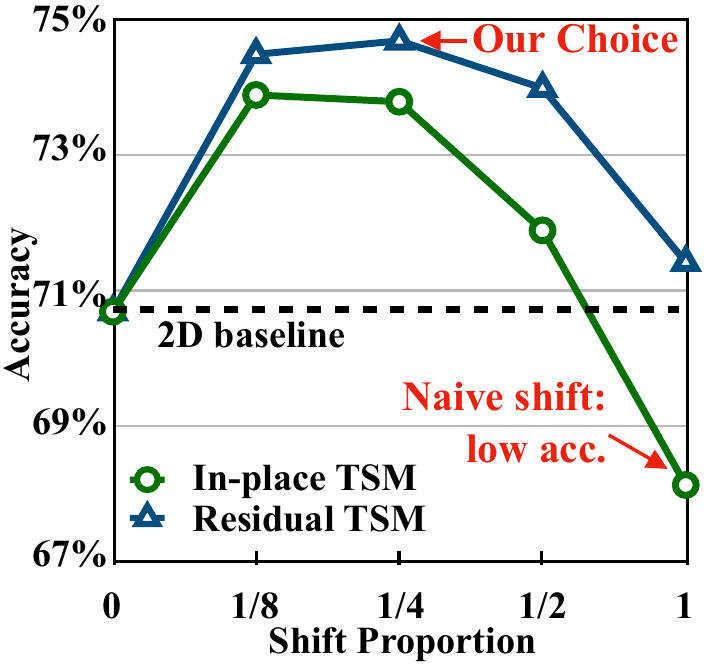}
	\caption{Residual \vs in-place.}
	\label{fig:tsm_discuss_b}
\end{subfigure}
\caption{\textbf{(a)} Latency overhead of TSM due to data movement. \textbf{(b)} Residual TSM achieve better performance than in-place shift. We choose 1/4 proportion residual shift  as our default setting. It achieves higher accuracy with a negligible overhead.}
\label{fig:tsm_discuss}
\end{figure}

To tackle the two problem from naive shift implementation, we discuss two technical contributions.


\subsubsection{Reducing Data Movement.}
To study the effect of data movement, we first measured the inference latency of TSM models and 2D baseline on different hardware devices. We shifted different proportion of the channels and measured the latency. We measured models with ResNet-50 backbone and 8-frame input using no shift (2D baseline), partial shift ($1/8, 1/4, 1/2$) and all shift (shift all the channels). The timing was measure on server GPU (NVIDIA Tesla P100), mobile GPU (NVIDIA Jetson TX2) and CPU (Intel Xeon E5-2690). We report the average latency from 1000 runs after 200 warm-up runs. We show the overhead of the shift operation as the percentage of the original 2D CNN inference time in~\ref{fig:tsm_discuss_a}. 
We observe the same overhead trend for different devices. If we shift all the channels, the latency overhead takes up to \textbf{13.7\%} of the inference time on CPU, which is definitely \textbf{non-negligible} during inference. On the other hand, if we only shift a small proportion of the channels, \eg, $1/8$, we can limit the latency overhead to  \textbf{only 3\%}. Therefore, we use \emph{partial shift} strategy in our TSM implementation to significantly bring down the memory movement cost.

\subsubsection{Keeping Spatial Feature Learning Capacity.}

\begin{figure}[t]
\centering
\begin{subfigure}[b]{0.22\textwidth}
	\includegraphics[width=\textwidth]{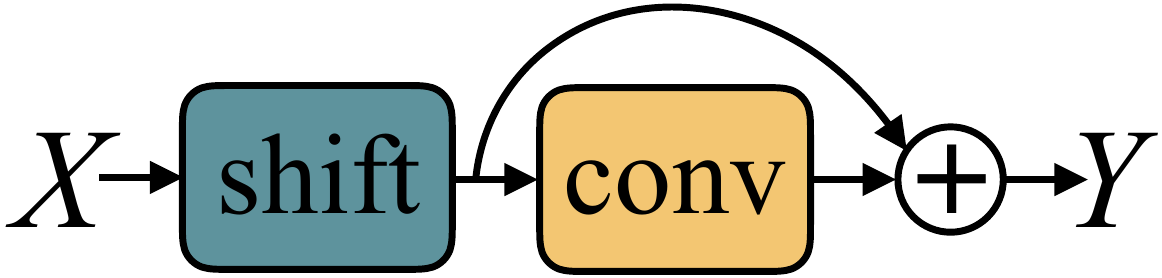}
	\caption{In-place TSM.}
	\label{fig:shift_inplace}
\end{subfigure}
~~
\begin{subfigure}[b]{0.22\textwidth}
	\includegraphics[width=\textwidth]{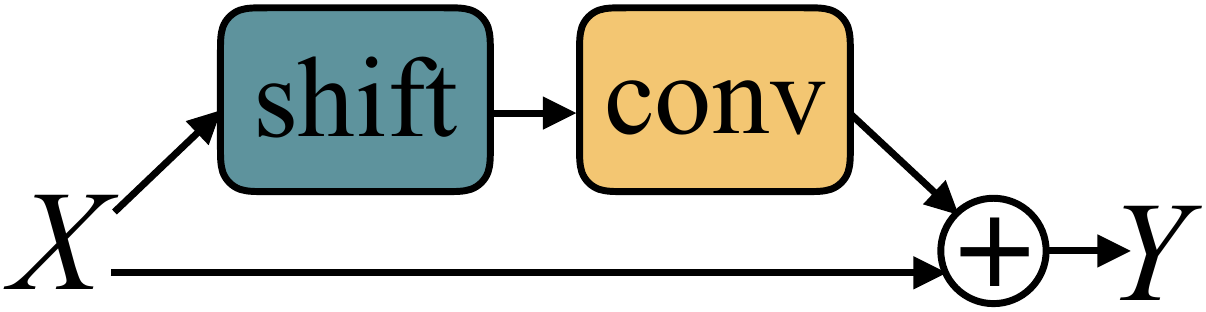}
	\caption{Residual TSM.}
	\label{fig:shift_residual}
\end{subfigure}
\caption{Residual shift is better than in-place shift. In-place shift happens before a convolution layer (or a residual block). Residual shift fuses temporal information inside a residual branch.}
\label{fig:shift_place}
\end{figure}

We need to balance the model capacity for spatial feature learning and temporal feature learning.
A straight-forward way to apply TSM is to insert it before each convolutional layer or residual block, as illustrated in Figure~\ref{fig:shift_inplace}. We call such implementation \emph{in-place shift}. It harms the spatial feature learning capability of the backbone model, especially when we shift a large amount of channels, since the information stored in the shifted channels is lost for the current frame.

To address such issue, we propose a variant of the shift module. Instead of inserting it in-place, we put the TSM \emph{inside} the residual branch in a residual block.
We denote such version of shift as \emph{residual shift} as shown in~\ref{fig:shift_residual}. Residual shift can address the degraded spatial feature learning problem, as all the information in the original activation is still accessible after temporal shift through identity mapping.

To verify our assumption, we compared the performance of in-place shift and residual shift on Kinetics~\cite{kay2017kinetics} dataset. We studied the experiments under different shift proportion setting. The results are shown in~\ref{fig:tsm_discuss_b}. We can see that residual shift achieves better performance than in-place shift for all shift proportion. Even we shift all the channels to neighboring frames, due to the shortcut connection, residual shift still achieves better performance than the 2D baseline.
Another finding is that the performance is related to the proportion of shifted channels: if the proportion is too small, the ability of temporal reasoning may not be enough to handle complicated temporal relationships; if too large, the spatial feature learning ability may be hurt. For residual shift, we found that the performance reaches the peak when $1/4$ ($1/8$ for each direction) of the channels are shifted. Therefore, we use this setting for the rest of the paper.

\section{TSM Video Network}


\subsection{Offline Models with Bi-directional TSM}
We insert bi-directional TSM to build offline video recognition models.
Given a video $V$, we first sample $T$ frames $F_i, F_1, ..., F_T$ from the video.
After frame sampling, 2D CNN baselines process each of the frames individually, and the output logits are averaged to give the final prediction. 
Our proposed \netFull model has exactly the same parameters and computation cost as 2D model. During the inference of convolution layers, the frames are still running independently just like the 2D CNNs. The difference is that  TSM is inserted for each residual block, which enables temporal information fusion at no computation.
For each inserted temporal shift module, the temporal receptive field will be enlarged by 2, as if running a convolution with the kernel size of 3 along the temporal dimension. Therefore, our \netFull model has a very large temporal receptive field to conduct highly complicated temporal modeling. In this paper, we used ResNet-50~\cite{he2016deep} as the backbone unless otherwise specified.

A unique advantage of TSM is that it can easily convert any off-the-shelf 2D CNN model into a pseudo-3D model that can handle both spatial and temporal information, without adding additional computation. Thus the deployment of our framework is hardware friendly: we only need to support the operations in 2D CNNs, which are already well-optimized at both framework level (CuDNN~\cite{chetlur2014cudnn}, MKL-DNN, TVM \cite{chen2018tvm}) and hardware level (CPU/GPU/TPU/FPGA).


\subsection{Online Models with Uni-directional TSM}

\begin{figure}[t]
\centering
\includegraphics[width=0.4\textwidth]{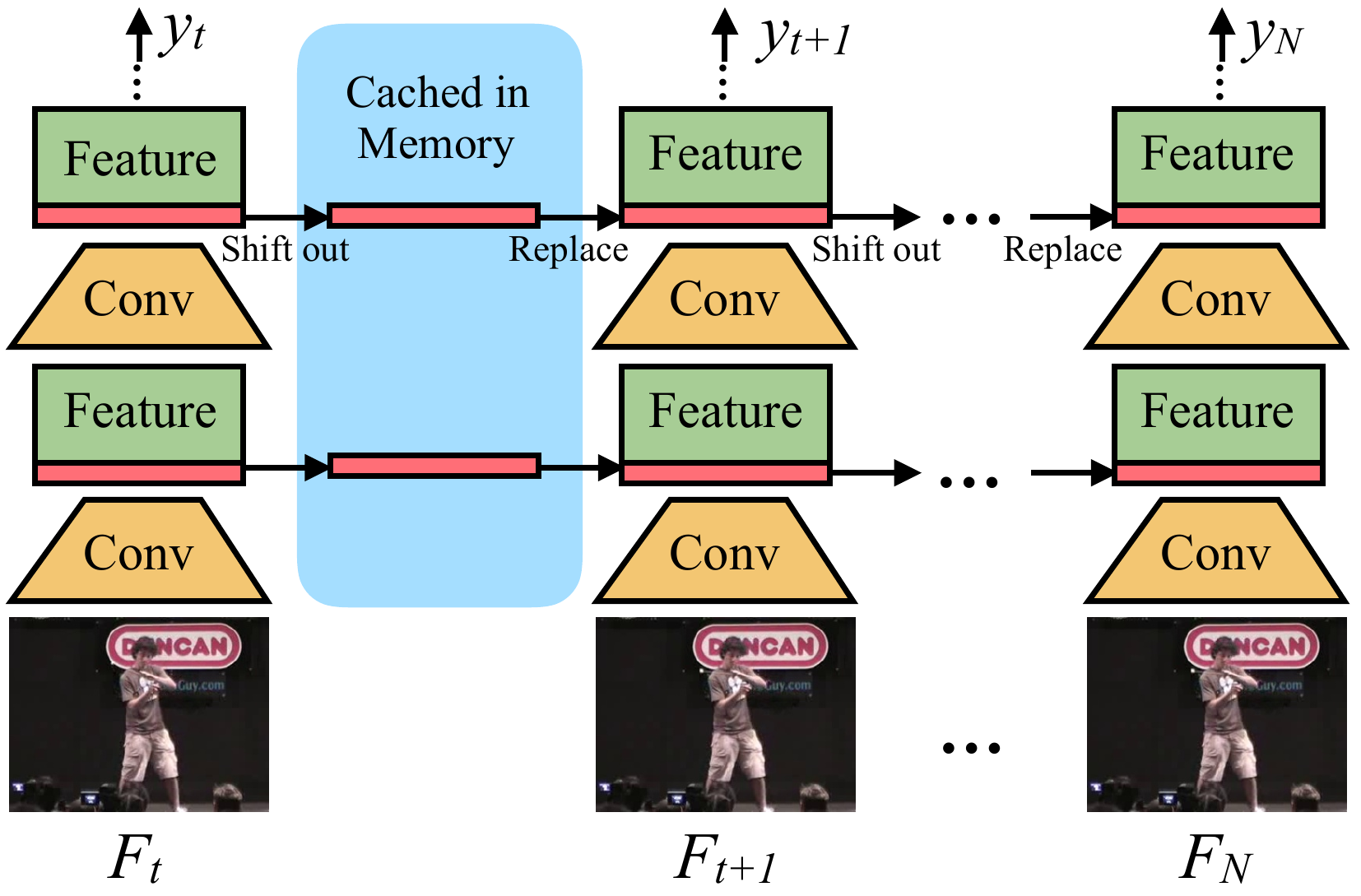}
\caption{Uni-directional TSM for online video recognition.}
\label{fig:singlesided}
\end{figure}

Video understanding from online video streams is important in real-life scenarios. Many real-time applications requires online video recognition with low latency, such as AR/VR and self-driving. In this section, we show that we can adapt TSM to achieve online video recognition while with multi-level temporal fusion.

As shown in Figure~\ref{fig:shift}, offline TSM shifts part of the channels bi-directionally, which requires features from future frames to replace the features in the current frame. If we only shift the feature from previous frames to current frames, we can achieve online recognition with uni-directional TSM.

The inference graph of uni-directional TSM for online video recognition is shown in Figure~\ref{fig:singlesided}. During inference, for each frame, we save the first 1/8 feature maps of each residual block and cache it in the memory. For the next frame, we replace the first 1/8 of the current feature maps with the cached feature maps. We use the combination of 7/8 current feature maps and 1/8 old feature maps to generate the next layer, and repeat. Using the uni-directional TSM for online video recognition shares several unique advantages: 

 \textbf{1. Low latency inference}. For each frame, we only need to replace and cache 1/8 of the features, without incurring any extra computations. Therefore, the latency of giving per-frame prediction is almost the same as the 2D CNN baseline. Existing methods like~\cite{zolfaghari2018eco} use multiple frames to give one prediction, which may leads to large latency.

\textbf{2. Low memory consumption}. Since we only cache a small portion of the features in the memory, the memory consumption is low. For ResNet-50, we only need 0.9MB memory cache to store the intermediate feature.

\textbf{3. Multi-level temporal fusion}. Most of the online method only enables late temporal fusion after feature extraction like~\cite{zhou2017temporal} or mid level temporal fusion~\cite{zolfaghari2018eco}, while our TSM enables all levels of temporal fusion. Through experiments (Table~\ref{tab:compare_something}) we find that multi-level temporal fusion is very important for complex temporal modeling. 

\section{Experiments}

We first show that TSM can significantly improve the performance of 2D CNN on video recognition while being computationally free and hardware efficient. It further demonstrated state-of-the-art performance on temporal-related datasets, arriving at a much better accuracy-computation pareto curve. TSM models achieve an order of magnitude speed up in measured GPU throughput compared to conventional I3D model from~\cite{wang2018videos}. Finally, we leverage uni-directional TSM to conduct low-latency and real-time online prediction on both video recognition and object detection.

\subsection{Setups}
\subsubsection{Training \& Testing.} 
We conducted experiments on video action recognition tasks.
The training parameters for the Kinetics dataset are: 100 training epochs, initial learning rate 0.01 (decays by 0.1 at epoch 40\&80), weight decay 1e-4, batch size 64, and dropout 0.5. For other datasets, we scale the training epochs by half.
For most of the datasets, the model is fine-tuned from ImageNet pre-trained weights; while HMDB-51~\cite{kuehne2011hmdb} and UCF-101~\cite{soomro2012ucf101} are too small and prone to over-fitting~\cite{wang2016temporal},
we followed the common practice~\cite{wang2016temporal, wang2017non} to 
fine-tune from Kinetics~\cite{kay2017kinetics} pre-trained weights and freeze the Batch Normalization~\cite{ioffe2015batch} layers.
For testing, when pursue high accuracy, we followed the common setting in~\cite{wang2017non, wang2018videos} to sample multiple clips per video (10 for Kinetics, 2 for others) and use the full resolution image with shorter side 256 for evaluation, so that we can give a direct comparison; when we consider the efficiency (\eg, as in Table~\ref{tab:compare_something}), we used just 1 clip per video and the center 224$\times$224 crop for evaluation. We keep the same protocol for the methods compared in the same table.



\subsubsection{Model.} 
To have an apple-to-apple comparison with the state-of-the-art method~\cite{wang2018videos}, we used the same backbone (ResNet-50) on the dataset ( Something-Something-V1~\cite{goyal2017something}).This dataset focuses on temporal modeling.  The difference is that \cite{wang2018videos} used 3D ResNet-50, while we used 2D ResNet-50 as the backbone to demonstrate efficiency.


\subsubsection{Datasets.}

Kinetics dataset~\cite{kay2017kinetics} is a large-scale action recognition dataset with 400 classes. As pointed in~\cite{zhou2017temporal, xie2018rethinking}, datasets like Something-Something (V1\&V2)~\cite{goyal2017something}, Charades~\cite{sigurdsson2016hollywood}, and Jester~\cite{jester}  are more focused on modeling the temporal relationships
, while UCF101~\cite{soomro2012ucf101}, HMDB51~\cite{kuehne2011hmdb}, and Kinetics~\cite{kay2017kinetics} are less sensitive to temporal relationships. Since TSM focuses on temporal modeling, we mainly focus on datasets with stronger temporal relationships like Something-Something. Nevertheless, we also observed strong results on the other datasets and reported it.

\subsection{Improving 2D CNN Baselines}

\renewcommand \arraystretch{0.9}
\begin{table}[t]
\small
\begin{center}
\begin{tabular}{cccccc}
\toprule
& \textbf{Dataset} &  \textbf{Model} &  \textbf{Acc1} &  \textbf{Acc5} &  \textbf{$\Delta$ Acc1}  \\ 
\toprule
\multirow{6}{*}{\rotatebox{90}{Less Temporal\quad}} & \multirow{2}{*}{Kinetics} &  TSN & 70.6 & 89.2 & \multirow{2}{*}{+3.5}  \\ 
 &  & Ours & \textbf{74.1} & \textbf{91.2} \\ \cmidrule(l){2-6}
& \multirow{2}{*}{UCF101}  & TSN & 91.7 &  99.2& \multirow{2}{*}{+4.2} \\ 
 &  & Ours & \textbf{95.9} & \textbf{99.7} \\ \cmidrule(l){2-6}
& \multirow{2}{*}{HMDB51}  & TSN & 64.7 & 89.9 & \multirow{2}{*}{+8.8} \\ 
 &  & Ours & \textbf{73.5} & \textbf{94.3} & \\ 
\midrule \midrule
\multirow{6}{*}{\rotatebox{90}{More Temporal\quad}}  & \multirow{2}{*}{\shortstack{Something\\V1}}  & TSN & 20.5  & 47.5 & \multirow{2}{*}{+28.0} \\ 
 &  & Ours & \textbf{47.3} & \textbf{76.2} \\ \cmidrule(l){2-6}
& \multirow{2}{*}{\shortstack{Something\\V2}}  &  TSN & 30.4 & 61.0 & \multirow{2}{*}{+31.3}\\ 
 &  & Ours & \textbf{61.7} & \textbf{87.4} \\ \cmidrule(l){2-6}
& \multirow{2}{*}{Jester}  & TSN & 83.9 & 99.6 & \multirow{2}{*}{+11.7} \\ 
 &  & Ours & \textbf{97.0} & \textbf{99.9} \\ 
\bottomrule
\end{tabular}
\end{center}
\caption{Our method consistently outperforms 2D counterparts on multiple datasets at zero extra computation (protocol: ResNet-50 8f input, 10 clips for Kinetics, 2 for others, full-resolution).}
\label{tab:compared_2d}
\end{table}

We can seamlessly inject TSM into a normal 2D CNN  and improve its performance on video recognition. In this section, we demonstrate a 2D CNN baseline can significantly benefit from TSM with double-digits accuracy improvement.  
We chose TSN~\cite{wang2016temporal} as the 2D CNN baseline. We used the same training and testing protocol for TSN and our \netFull. The only difference is with or without TSM.  

\subsubsection{Comparing Different Datasets. } We compare the results on several action recognition datasets in Table~\ref{tab:compared_2d}. The chart is split into two parts. The upper part contains datasets Kinetics~\cite{kay2017kinetics}, UCF101~\cite{soomro2012ucf101}, HMDB51~\cite{kuehne2011hmdb}, where temporal relationships are less important, while our \netFull still consistently outperforms the 2D TSN baseline at no extra computation. For the lower part, we present the results on Something-Something V1 and V2~\cite{goyal2017something} and Jester~\cite{jester}, which depend heavily on temporal relationships. 2D CNN baseline cannot achieve a good accuracy, but once equipped with TSM, the performance improved by double digits. 


\subsubsection{Scaling over Backbones.} TSM scales well to backbones of different sizes. We show the Kinetics top-1 accuracy with MobileNet-V2~\cite{sandler2018mobilenetv2}, ResNet-50~\cite{he2016deep}, ResNext-101~\cite{xie2017aggregated} and ResNet-50 + Non-local module~\cite{wang2017non} backbones in Table~\ref{tab:compared_backbone}. TSM consistently improves the accuracy over different backbones, even for NL R-50, which already has temporal modeling ability.

\subsection{Comparison with State-of-the-Arts}

\renewcommand \arraystretch{1.}
\begin{table*}[t]
\setlength{\tabcolsep}{5pt}
\small
\begin{center}
\begin{tabular}{cccccccc}
\toprule
\textbf{Model}
& \textbf{Backbone} & \textbf{\#Frame} & \textbf{FLOPs/Video} & \textbf{\#Param.} & \textbf{Val Top-1} & 
\textbf{Val Top-5} & \textbf{Test Top-1}  \\ 
\toprule
TSN~\cite{zhou2017temporal} & BNInception & 8  &16G & 10.7M & 19.5 & - & - \\ 
TSN (our impl.) & ResNet-50 & 8  & 33G & 24.3M & 19.7 & 46.6 & -   \\ 
TRN-Multiscale~\cite{zhou2017temporal} & BNInception & 8 &16G & 18.3M & 34.4 & - & 33.6 \\ 
TRN-Multiscale (our impl.) & ResNet-50 & 8 &33G & 31.8M & 38.9 & 68.1  & - \\ 
Two-stream TRN\textsubscript{RGB+Flow}~\cite{zhou2017temporal} & BNInception & 8+8 & - & 36.6M & 42.0 & - & 40.7 \\ \midrule
ECO~\cite{zolfaghari2018eco} & BNIncep+3D Res18  & 8  & 32G & 47.5M & 39.6 & - & - \\
ECO~\cite{zolfaghari2018eco} & BNIncep+3D Res18 & 16 & 64G & 47.5M & 41.4 & - & - \\ 
ECO\textsubscript{\emph{En}}\emph{Lite}~\cite{zolfaghari2018eco} & BNIncep+3D Res18 & 92 & 267G & 150M & 46.4 & - & 42.3   \\
ECO\textsubscript{\emph{En}}\emph{Lite}\textsubscript{RGB+Flow}~\cite{zolfaghari2018eco} & BNIncep+3D Res18 & 92+92 & - & 300M & 49.5 & - & 43.9 \\ \midrule
I3D from~\cite{wang2018videos} & 3D ResNet-50 & 32$\times$2clip & 153G\tablefootnote{\label{note1} We reported the performance of NL I3D described in~\cite{wang2018videos}, which is a variant of the original NL I3D~\cite{wang2017non}. It uses fewer temporal dimension pooling to achieve good performance, but also incur larger computation.}$\times$2  & 28.0M & 41.6 & 72.2 & - \\ 
Non-local I3D from~\cite{wang2018videos} & 3D ResNet-50 & 32$\times$2clip & 168G\footnoteref{note1}$\times$2 & 35.3M & 44.4 & 76.0 & - \\  
Non-local I3D + GCN~\cite{wang2018videos} & 3D ResNet-50+GCN & 32$\times$2clip & 303G\tablefootnote{\label{note2}Includes parameters and FLOPs of the Region Proposal Network.}$\times$2 & 62.2M\footnoteref{note2} & 46.1 & 76.8  & 45.0\\ 
\midrule
CorrNet-50~\cite{wang2020video} & R(2+1)D-50 & 32$\times$10clip & 115G$\times$10 & - & 49.3 & - & -\\
ip-CSN-152~\cite{tran2019video} & 3D ResNet-152 & 32$\times$10clip & 83.3G$\times$10 & 33.0M &53.3 &- &- \\
\midrule\midrule
\netFull & ResNet-50 & 8 & 33G & 24.3M & 45.6 & 74.2 & - \\
\netFull & ResNet-50 & 16 & 65G & 24.3M & 47.2 & 77.1 & 46.0 \\ \midrule
\netHead\textsubscript{\emph{En}} & ResNet-50 & 24 & 98G & 48.6M & 49.7 & 78.5 & - \\
\netHead\textsubscript{RGB+Flow} & ResNet-50 & 16+16 & - & 48.6M & \textbf{52.6} & \textbf{81.9}  & \textbf{50.7}\\
\bottomrule
\end{tabular}
\end{center}
\caption{Comparing \netFull against other methods on Something-Something dataset (center crop, 1 clip/video unless otherwise specified).  }
\label{tab:compare_something}
\end{table*}

\renewcommand \arraystretch{0.9}
\begin{table}[t]
\small
\begin{center}
\begin{tabular}{ccccc}
\toprule
& \textbf{Mb-V2} &  \textbf{R-50} &  \textbf{RX-101} &  \textbf{NL R-50}\\ 
\midrule
TSN & 66.5 & 70.7 & 72.4 & 74.6 \\
TSM & 69.5 & 74.1 & 76.3 & 75.7 \\
\midrule
$\Delta$Acc. & +3.0 & +3.4 & +3.9 & +1.1  \\
\bottomrule 
\end{tabular}
\end{center}
\caption{TSM can consistently improve the performance over different backbones on Kinetics dataset.}
\label{tab:compared_backbone}
\end{table}


\netFull not only significantly improves the 2D baseline but also outperforms state-of-the-art methods, which heavily rely on 3D convolutions. We compared the performance of our offline (bi-directional) TSM model with state-of-the-art methods on both Something-Something V1\&V2 because these two datasets focus on temporal modeling.

\subsubsection{Something-Something-V1.} Something-Something-V1 is a challenging dataset, as activity cannot be inferred merely from individual frames (\eg, pushing something from \emph{right to left}).
We compared TSM with current state-of-the-art methods in Table~\ref{tab:compare_something}. 
We only applied center crop during testing to ensure the efficiency unless otherwise specified. TSM achieves \emph{the first place} on the leaderboard upon publication.

We first show the results of the 2D based methods TSN~\cite{wang2016temporal} and TRN~\cite{zhou2017temporal}. TSN with different backbones fails to achieve decent performance ($<$20\%  Top-1) due to the lack of temporal modeling. For TRN, although \textbf{late temporal fusion} is added after feature extraction, the performance is still significantly lower than state-of-the-art methods, showing the importance of temporal fusion across all levels. 

The second section shows the state-of-the-art efficient video understanding framework ECO~\cite{zolfaghari2018eco}. 
ECO uses an early 2D + late 3D architecture which enables \textbf{medium-level temporal fusion}.
Compared to ECO, our method achieves better performance at a smaller FLOPs. For example, when using 8 frames as input, our \netFull achieves 45.6\% top-1 accuracy with 33G FLOPs, which is 4.2\% higher accuracy than ECO with 1.9$\times$ less computation. The ensemble versions of ECO (ECO\textsubscript{\emph{En}}\emph{Lite} and ECO\textsubscript{\emph{En}}\emph{Lite}\textsubscript{RGB+Flow}, using an ensemble of  \{16, 20, 24, 32\} frames as input) did achieve competitive results, but the computation and parameters are too large for deployment. While our model is much more efficient: we only used \{8, 16\} frames model for ensemble (\netHead\textsubscript{\emph{En}}), and the model achieves better performance using 2.7$\times$ less computation and 3.1$\times$ fewer  parameters.

The third section contains previous state-of-the-art methods: Non-local I3D + GCN~\cite{wang2018videos}, that enables \textbf{all-level temporal fusion}. The GCN needs a Region Proposal Network~\cite{ren2015faster} trained on MSCOCO object detection dataset~\cite{lin2014microsoft} to generate the bounding boxes, which is unfair to compare since external data (MSCOCO) and extra training cost is introduced. Thus we compared \netFull to its CNN part: Non-local I3D. Our \netFull (8f) achieves 1.2\% better accuracy with $10\times$ fewer FLOPs on the validation set compared to the Non-local I3D network. Note that techniques like Non-local module~\cite{wang2017non} are orthogonal to our work, which could also be added to our framework to boost the performance further.

We further include two recent state-of-the-art methods that achieve state-of-the-art performance: CorrNet~\cite{wang2020video} and CSN~\cite{tran2019video}. For CorrNet, we compare to CorrNet-50 which has a similar backbone shape; For CSN, we compare to ip-CSN-152, which is the largest model and achieves the highest accuracy. Both methods achieve high accuracy on Something-Something dataset. However, they still require sampling 10 clips to get the average prediction. The total computation is larger than 800G FLOPs, which is not practical for edge deployment.

\subsubsection{Generalize to Other Modalities.} We also show that our proposed method can generalize to other modalities like optical flow. To extract the optical flow information between frames, we followed~\cite{wang2016temporal} to use the TVL1 optical flow algorithm~\cite{zach2007duality} implemented in OpenCV with CUDA. We conducted two-stream experiments on both Something-Something V1 and V2 datasets, and it consistently improves over the RGB performance: introducing optical flow branch brings 5.4\% and 2.6\% top-1 improvement on V1 and V2.

\renewcommand \arraystretch{0.95}
\begin{table}[t]
\setlength{\tabcolsep}{7pt}
\small
\begin{center}
\begin{tabular}{ccccc}
\toprule
\multirow{2}{*}{\textbf{Method}} & \multicolumn{2}{c}{\textbf{Val}}   &  \multicolumn{2}{c}{\textbf{Test}} \\
\cmidrule(lr){2-3}\cmidrule(lr){4-5}
& \textbf{Top-1} & \textbf{Top-5}   &  \textbf{Top-1} & \textbf{Top-5} \\ \midrule
TSN (our impl.) & 30.0 & 60.5 & - & - \\
MultiScale TRN~\cite{zhou2017temporal} & 48.8 & 77.6 & 50.9 & 79.3 \\
2-Stream TRN~\cite{zhou2017temporal} & 55.5 & 83.1 & 56.2 & 83.2 \\
\midrule
\netHead\textsubscript{8F} & 59.1 & 85.6 & - & - \\
\netHead\textsubscript{16F} & 63.4 & 88.5  & 64.3 & 89.6 \\
\netHead\textsubscript{RGB+Flow} & \textbf{66.0} & \textbf{90.5} & \textbf{66.6} & \textbf{91.3}\\ 
\bottomrule
\end{tabular}
\end{center}
\caption{Results on Something-Something-V2. Our \netFull achieves state-of-the-art performance.}
\label{tab:somethingv2}
\end{table}

\begin{figure}[t]
\centering
\includegraphics[width=0.4\textwidth]{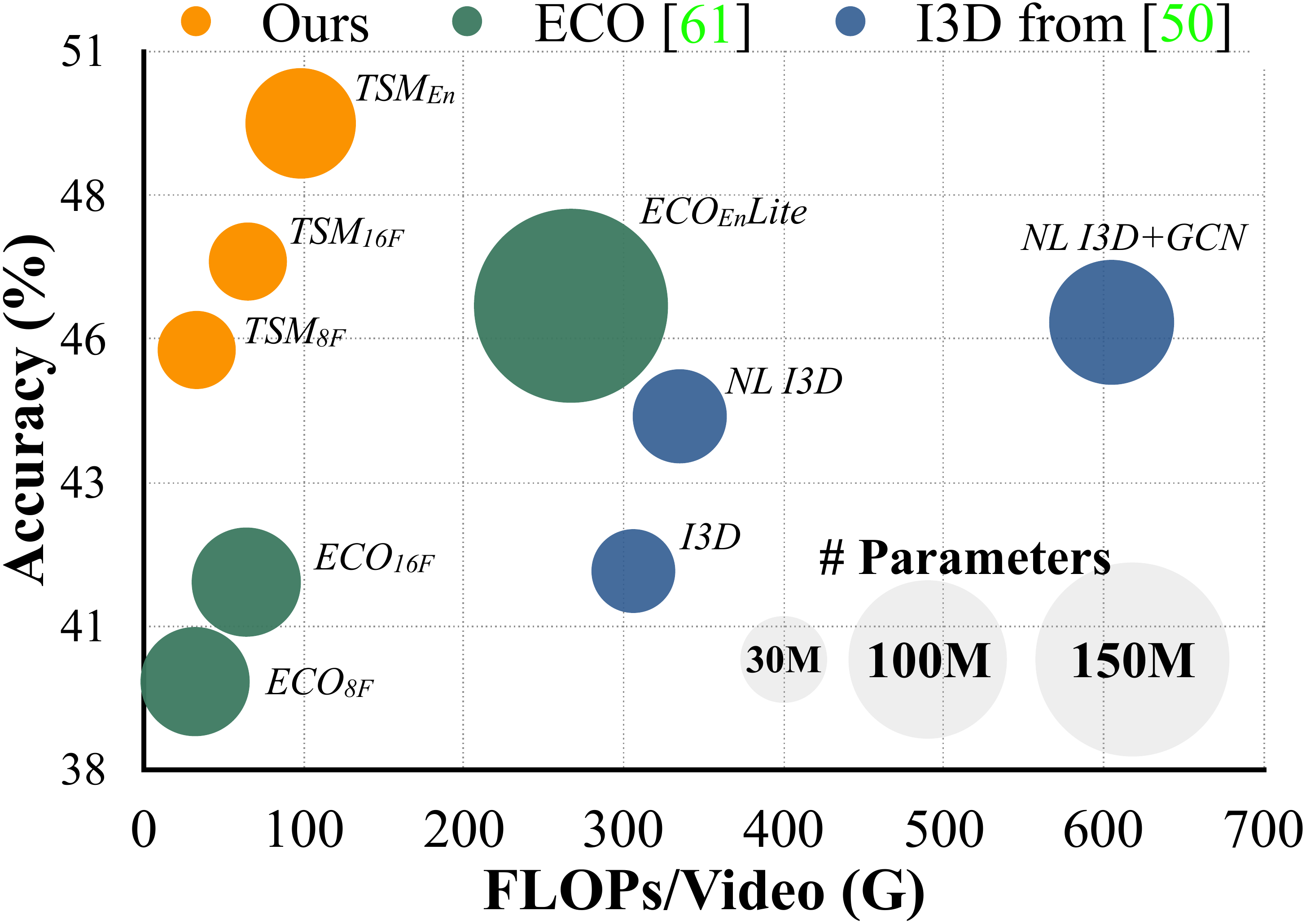}
\caption{
TSM enjoys better accuracy-cost trade-off than I3D family and ECO family on Something-Something-V1~\cite{goyal2017something} dataset.
(GCN includes the cost of ResNet-50 RPN to generate region proposals.)}
\label{fig:acc_vs_flops}
\end{figure}

\subsubsection{Something-Something-V2.} We also show the result on Something-Something-V2 dataset, which is a newer release to its previous version. The results compared to other state-of-the-art methods are shown in Table~\ref{tab:somethingv2}. 
On Something-Something-V2 dataset, we achieved state-of-the-art performance while only using RGB input. 


\subsubsection{Cost \vs Accuracy.}

Our \netFull model achieves very competitive performance while enjoying high efficiency and low computation cost for fast inference. We show the FLOPs for each model in Table~\ref{tab:compare_something}. 
Although GCN itself is light, the method used a ResNet-50 based Region Proposal Network~\cite{ren2015faster} to extract bounding boxes, whose cost is also considered in the chart.
Note that the computation cost of optical flow extraction is usually larger than the video recognition model itself. Therefore, we do not report the FLOPs of two-stream based methods.

We show the accuracy, FLOPs, and number of parameters trade-off in Figure~\ref{fig:acc_vs_flops}. The accuracy is tested on the validation set of Something-Something-V1 dataset, and the number of parameters is indicated by the area of the circles. We can see that our \netFull based methods have a better Pareto curve than both previous state-of-the-art efficient models (ECO based models) and high-performance models (non-local I3D based models). 
\netFull models are both efficient and accurate. It can achieve state-of-the-art accuracy at high efficiency: it achieves better performance while consuming $3\times$ less computation than the ECO family
. Considering that ECO is already an efficiency-oriented design, our method enjoys highly competitive hardware efficiency.

\subsubsection{Kinetics.}

\begin{table}[t]
\small
\begin{center}
\begin{tabular}{lcc}
\toprule
\textbf{Model} &  \textbf{Latency/Clip} &   \textbf{Top-1}\\ 
\midrule
I3D NL R-50~\cite{wang2017non} & 30.7ms & 74.9\%\\
SlowFast R-50~\cite{feichtenhofer2019slowfast} & 33.3ms & 75.6\%\\ 
\textbf{TSM NL R-50} & \textbf{25.6ms} & \textbf{75.7\%} \\ 
\midrule
X3D-M~\cite{feichtenhofer2020x3d} & 73.5ms & 76.0\% \\
ir-CSN-152~\cite{tran2019video} & 138.6ms & \textbf{76.8\%}\\
\textbf{TSM RX-101} & \textbf{40.7ms}   & 76.3\%\\
\bottomrule 
\end{tabular}
\end{center}
\vspace{-10pt}
\caption{Compare to state-of-the-art methods on Kinetics. TSM can achieve higher or comparable  performance at a lower inference latency.}
\label{tab:compared_kinetics}
\end{table}

Although Kinetics does not focus on temporal modeling (Table~\ref{tab:compared_2d}), we compare TSM with state-of-the-art methods on Kinetics to give a comprehensive comparison. The results are show in Table~\ref{tab:compared_kinetics}. 

We compare to several state-of-the-art methods: Non-local networks~\cite{wang2017non} (I3D NL R-50), SlowFast~\cite{feichtenhofer2019slowfast} (SlowFast R-50), X3D~\cite{feichtenhofer2020x3d} (X3D-M), and CSN~\cite{tran2019video} (ir-CSN-152). We report the latency and accuracy trade-off of different methods. The latency is measured on NVIDIA RTX 2080 Ti GPU using batch size 1. We first warm-up the inference for 100 iterations and measure the average latency of the next 200 iterations. TSM can achieve higher or comparable performance at a lower inference latency. TSM (TSM NL R-50) achieves same-level of accuracy compared to SlowFast network (SlowFast R-50) at 1.3$\times$ lower latency. TSM (TSM RX-101) also outperforms X3D (X3D-M) at 1.8$\times$ lower latency. Notice that though X3D has a small computation FLOPs, its inferior hardware efficiency leads to the slow inference speed. 
ir-CSN-152 achieves slightly higher accuracy than TSM, but TSM runs 3.4$\times$ faster. TSM is very competitive for accuracy-speed trade-off.


\subsection{Latency and Throughput Speedup}

\renewcommand \arraystretch{1.2}
\begin{table}[t]
\setlength{\tabcolsep}{2.5pt}
\begin{center}
\footnotesize{
    \begin{tabular}{ccccccc}
    \toprule
    \multirow{2}{*}{\textbf{Model}} &
     \multicolumn{4}{c}{\textbf{Efficiency Statistics}}  & \multicolumn{2}{c}{\textbf{Accuracy}} \\ 
     \cmidrule(lr){2-5}\cmidrule(lr){6-7}
     &  \textbf{FLOPs} & \textbf{Param.} &  \textbf{Latency} & \textbf{Thrput.} & \textbf{Sth.} & \textbf{Kinetics} \\ \toprule
    I3D from~\cite{wang2018videos} & 306G & 35.3M & 165.3ms & 6.1V/s & 41.6\% & -\\  
    ECO\textsubscript{16F}~\cite{zolfaghari2018eco}& 64G & 47.5M  & 30.6ms & 45.6V/s & 41.4\% & - \\ 
    \midrule
    I3D from~\cite{wang2017non} & \textbf{33G} & 29.3M & 25.8ms & 42.4V/s & - & 73.3\% \\
    I3D\textsubscript{replace} & 48G & 33.0M & 28.0ms & 37.9V/s & 44.9\% & -\\
    \midrule
     \netHead\textsubscript{8F}  & \textbf{33G} & \textbf{24.3M} & \textbf{17.4ms} & \textbf{77.4V/s} & 45.6\% & 74.1\%  \\ 
    \netHead\textsubscript{16F}  & 65G & 24.3M & 29.0ms & 39.5V/s & \textbf{47.2\%} & \textbf{74.7\%} \\ 
    \bottomrule
    \end{tabular}
}
\end{center}
\caption{TSM enjoys low GPU inference latency and high throughput.
V/s means videos per second, higher the better (Measured on NVIDIA Tesla P100 GPU). }
\label{tab:runtime}
\end{table}


The measured inference latency and throughput are important for the large-scale video understanding.
TSM has low latency and high throughput. We performed measurement on a single NVIDIA Tesla P100 GPU. We used batch size of 1 for latency measurement; batch size of16 for throughput measurement. 
We made two comparisons: 

(1) Compared with the I3D model from~\cite{wang2018videos}, our method is faster by an order of magnitude
at 1.8\% higher accuracy (Table~\ref{tab:runtime}). We also compared our method to the state-of-the-art efficient model ECO~\cite{zolfaghari2018eco}: Our \netFull model has $1.75\times$ lower latency (17.4ms \vs 30.6ms), $1.7\times$ higher throughput, and achieves 2\% better accuracy.  ECO has a two-branch (2D+3D) architecture, while TSM only needs the in-expensive 2D backbone.

(2) We then compared TSM to efficient 3D model designs.
One way is to only inflate the first $1\times1$ convolution in each of the block as in~\cite{wang2017non}, denoted as "I3D from~\cite{wang2017non}" in the table. 
Although the FLOPs are similiar due to pooling, it suffers from 1.5$\times$ higher latency and only 55\% the throughput compared with TSM, with worse accuracy. 
We speculate the reason is that TSM model only uses 2D convolution which is highly optimized for hardware.
To exclude the factors of backbone design, we replace every TSM primitive with $3\times1\times1$ convolution and denote this model as I3D\textsubscript{replace}. It is still much slower than TSM and performs worse.


\subsection{Online Recognition with TSM}

\begin{figure*}[!h]
\centering
\includegraphics[width=\textwidth]{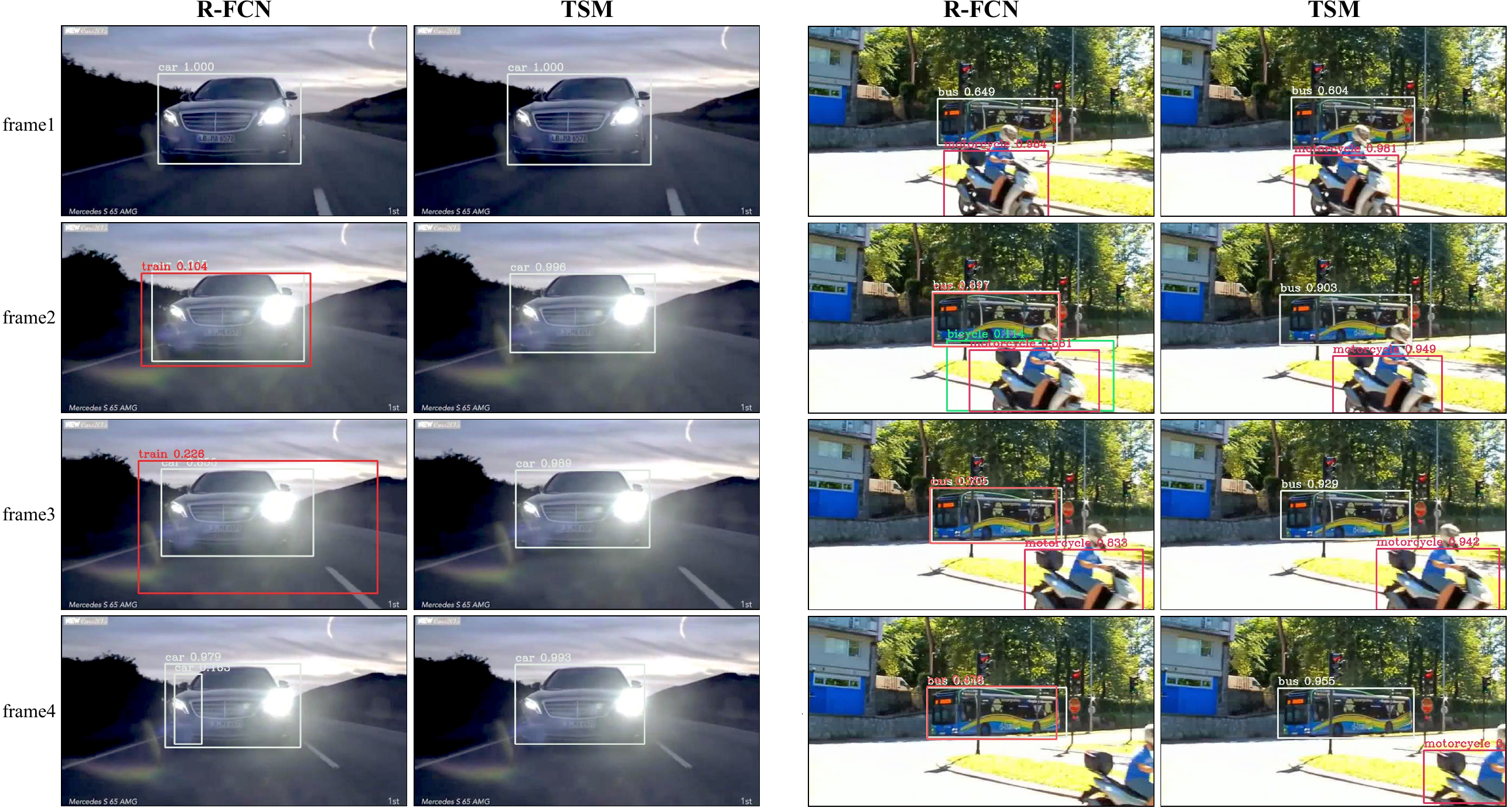}
\caption{TSM improves detection results with the help of temporal cues. For the left video, 2D baseline R-FCN generates false positive due to the glare of car headlight on frame 2/3/4, while TSM does not have such issue by considering the temporal information. For the right video, R-FCN generates false positive surrounding the bus due to occlusion by the traffic sign on frame 2/3/4. Also, it fails to detect motorcycle on frame 4 due to occlusion. TSM model addresses such issues with the help of temporal information. }
\label{fig:det_vis}
\end{figure*}


\renewcommand \arraystretch{1.}
\begin{table}[t]
\setlength{\tabcolsep}{3pt}
\small
\begin{center}
\begin{tabular}{cccccc}
\toprule
\textbf{Model} & \textbf{Latency}
& \textbf{Kinetics} & \textbf{UCF101} & \textbf{HMDB51} & \textbf{Something}  \\ 
\toprule
TSN & 4.7ms & 70.6\%  & 91.7\% & 64.7\% & 20.5\% \\ \midrule
+Offline & - & 74.1\% &  95.9\% & 73.5\% & 47.3\% \\ 
+Online & 4.8ms & 74.3\% & 95.5\% & 73.6\%  & 46.3\% \\
\bottomrule
\end{tabular}
\end{center}
\caption{Comparing the accuracy of offline TSM and online TSM on different datasets. Online TSM brings negligible latency overhead.}
\label{tab:offline_online}
\end{table}

\begin{figure}[t]
\centering
\includegraphics[width=0.38\textwidth]{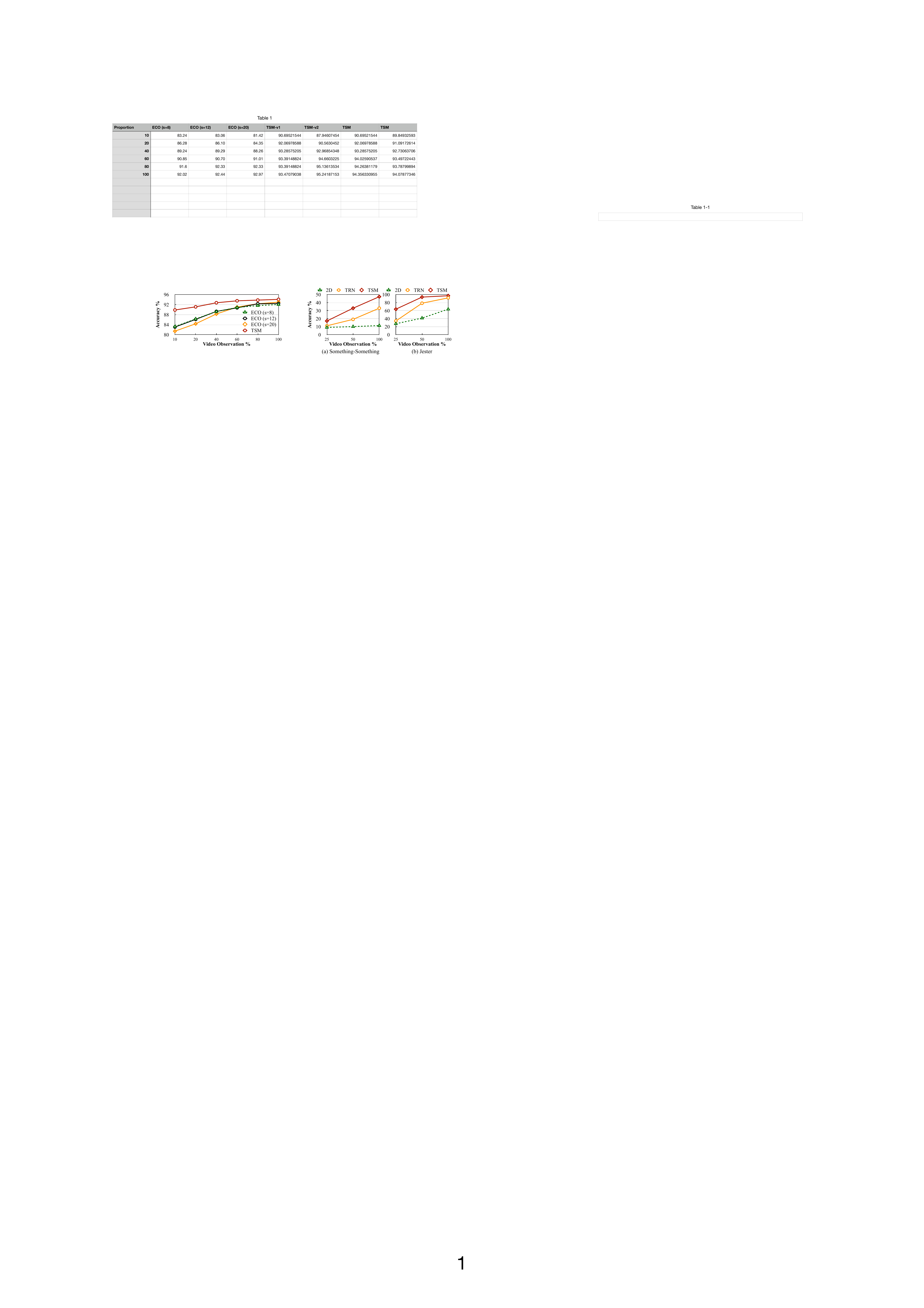}
\caption{Early recognition on UCF101.  TSM gives high prediction accuracy after only observing a small portion of the video.}
\label{fig:early_recognition}
\end{figure}

\begin{figure}[t]
\centering
\vspace{-10pt}
\includegraphics[width=0.43\textwidth]{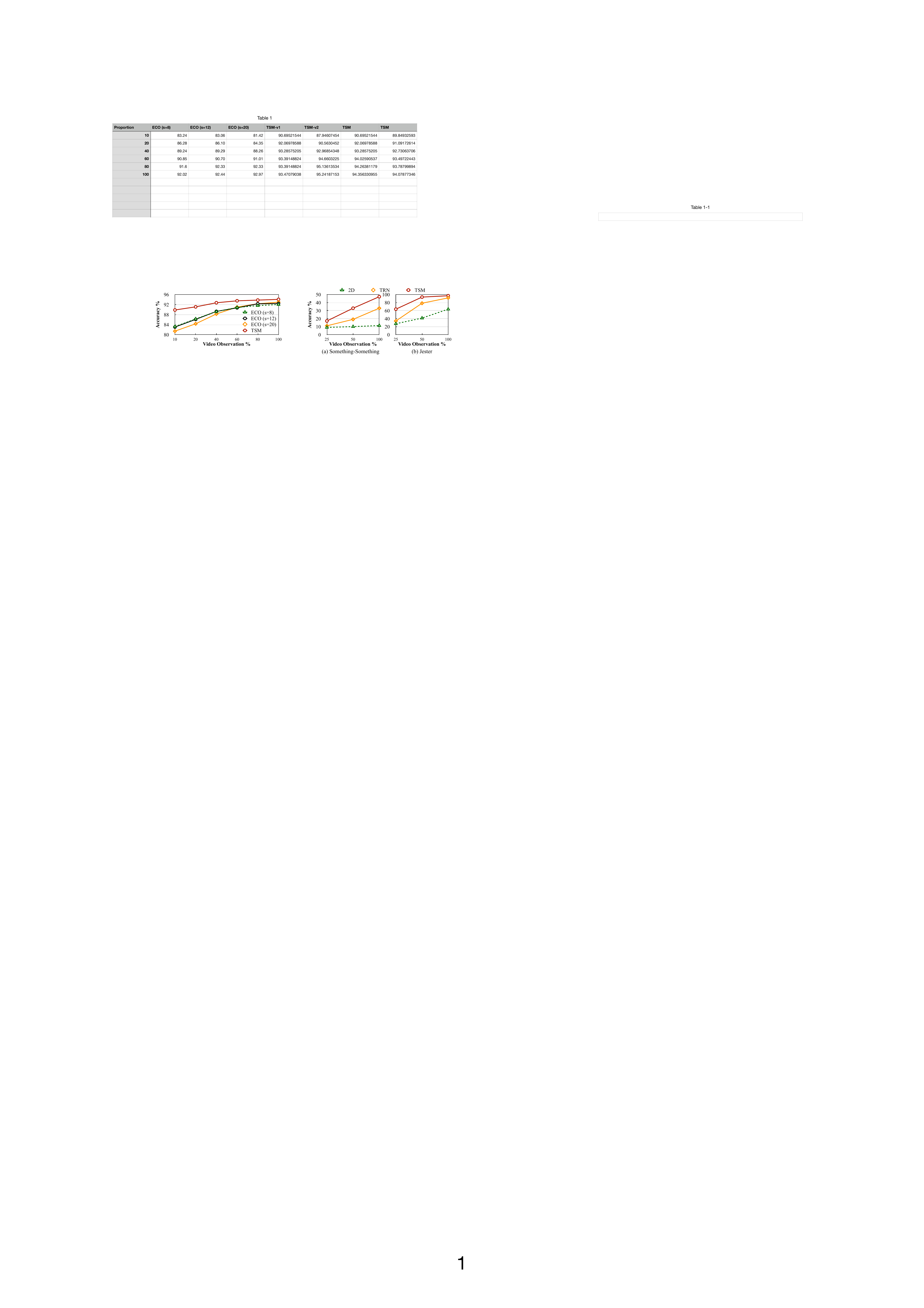}
\vspace{-5pt}
\caption{Early recognition on Something-Something and Jester datasets. TSM consistently outperforms TRN~\cite{zhou2017temporal} at various portions by a large margin.}
\label{fig:early_recognition2}
\end{figure}


\subsubsection{Online \vs Offline}
Online TSM models shift the feature maps uni-directionally so that it can give predictions in real time. 
We compare the performance of offline and online TSM models to show that online TSM can still achieve comparable performance. 
Follow~\cite{zolfaghari2018eco}, we use the prediction averaged from all the frames to compare with offline models, \ie, we compare the performance after observing the whole videos. 
The performance is provided in Table~\ref{tab:offline_online}.
We can see that for less temporal related datasets like Kinetics, UCF101 and HMDB51, the online models achieve comparable and sometimes even better performance compared to the offline models. While for more temporal related datasets Something-Something, online model performs worse than offline model by 1.0\%. Nevertheless, the performance of online model is still significantly better than the 2D baseline.

We also compare the per-frame prediction latency of pure 2D backbone (TSN) and our online TSM model. We compile both models with TVM~\cite{chen2018tvm} on GPU.
Our online TSM model only adds to less than 0.1ms latency overhead per frame while bringing up to 25\% accuracy improvement. It demonstrates online TSM is hardware-efficient for latency-critical real-time applications.

\subsubsection{Early Recognition}

Early recognition aims to classify the video while only observing a small portion of the frames. It gives fast response to the input video stream. Here we compare the early video recognition performance with ECO~\cite{zolfaghari2018eco} on UCF101  (Figure~\ref{fig:early_recognition}) and TRN~\cite{zhou2017temporal} on Something-Something and Jester (Figure~\ref{fig:early_recognition2}). Compared to ECO, TSM gives much higher accuracy, especially when only observing a small portion of the frames. For example, when only observing the first 10\% of video frames, TSM model can achieve 90\% accuracy, which is 6.6\% higher than the best ECO model. TSM also consistently outperforms TRN at various observation percentages by a large margin.

\subsubsection{Edge Deployment}

\renewcommand \arraystretch{1}
\begin{table}[t]
\setlength{\tabcolsep}{3pt}
\small
\begin{center}
\begin{tabular}{cccccccc}
\toprule
\multirow{2}{*}{\textbf{Devices}} & \multicolumn{2}{c}{Jetson Nano} & \multicolumn{2}{c}{Jetson TX2} & \multirow{2}{*}{Rasp.} &   \multirow{2}{*}{Note8} & \multirow{2}{*}{Pixel1}  \\ \cmidrule(lr){2-3} \cmidrule(lr){4-5}
& CPU & GPU & CPU & GPU& \\
\toprule
\textbf{FPS} & 20.9 & 74.6 & 27.5 & 117.6 & 14.4  & 29.0 & 21.1 \\
\textbf{Power} (watt) & 4.8  & 4.5 & 5.6 & 5.8 & 3.8 & - & - \\
\bottomrule
\end{tabular}
\end{center}
\caption{TSM efficiently runs on edge devices with low latency.}
\label{tab:edge_deploy}
\end{table}

TSM is mobile device friendly. 
We build an online TSM model with MobileNet-V2 backbone, which achieves 69.5\% accuracy on Kinetics. The latency and energy on NVIDIA Jetson Nano \& TX2, Raspberry Pi 4B, Samsung Galaxy Note8, Google Pixel-1 is shown in Table~\ref{tab:edge_deploy}. The models are compiled using TVM~\cite{chen2018tvm}. Power is measured with a power meter, subtracting the static power. TSM achieves low latency and low power on edge devices.

\section{Online Object Detection}



Real-time online video object detection is an important application in self-driving vehicles, robotics, \etc.
Most exiting methods treat video detection as image detection per frame, which is not robust since temporal information is not considered. Other methods on video object detection~\cite{zhu2017flow} fuses information along temporal dimension after the 2D feature is extracted by the backbone, which results in high latency, and also loses low-level temporal cues.

Here we show that we can enable temporal fusion in online video object detection by injecting our uni-directional TSM into the backbone. We show that we can significantly improve the performance of video detection by simply modifying the backbone with online TSM, without changing the detection module design or using optical flow features.

We conducted experiments with R-FCN~\cite{dai16rfcn} detector on ImageNet-VID~\cite{russakovsky2015imagenet} dataset. Following the setting in~\cite{zhu2017flow}, we used ResNet-101~\cite{he2016deep} as the backbone for R-FCN detector. For TSM experiments, we inserted uni-directional TSM to the backbone, while keeping other settings the same. We used the official training code of~\cite{zhu2017flow} to conduct the experiments, and the results are shown in Table~\ref{tab:detection}. 
Compared to 2D baseline R-FCN~\cite{dai16rfcn}, our online TSM model significantly improves the performance, especially on the fast moving objects, where TSM increases mAP by $4.6\%$. 
FGFA~\cite{zhu2017flow} is a strong baseline that uses optical flow to aggregate the temporal information from 21 frames (past 10 frames and future 10 frames) for offline video detection. Compared to FGFA, TSM can achieve similar or higher performance while enabling online recognition (using information from only past frames) at much smaller latency per frame.
The latency overhead of TSM module itself is less than 1ms per frame, making it a practical tool for real deployment.

We visualize the detection results of two video clips in Figure~\ref{fig:det_vis}. In the left video clip, 2D baseline R-FCN generates false positive due to the glare of car headlight on frame 2/3/4, while TSM suppresses false positive. In the right video clip, R-FCN generates false positive surrounding the bus due to occlusion by the traffic sign on frame 2/3/4. Also, it fails to detect motorcycle on frame 4 due to occlusion. TSM model addresses such issues with the help of temporal information. 

\renewcommand \arraystretch{0.9}
\begin{table}[t]
\setlength{\tabcolsep}{2.5pt}
\begin{center}
\footnotesize{
    \begin{tabular}{cccccccc}
    \toprule
    \multirow{2}{*}{\textbf{Model}} & \multirow{2}{*}{\textbf{Online}} & \multirow{2}{*}{\textbf{\shortstack{Need\\Flow}}} & \multirow{2}{*}{\textbf{Latency}} &
    \multicolumn{4}{c}{\textbf{mAP}} \\ 
    \cmidrule(lr){5-8}
    & & & & \textbf{Overall} & \textbf{Slow} & \textbf{Medium} & \textbf{Fast} \\
    \midrule
    R-FCN~\cite{dai16rfcn} & \checkmark & & 1$\times$ & 74.7 & 83.6 & 72.5 & 51.4\\
    FGFA~\cite{zhu2017flow} & & \checkmark & 2.5$\times$  & 75.9 & \textbf{84.0} & 74.4 & 55.6  \\
    \midrule
    Online TSM & \checkmark &  & 1$\times$ & \textbf{76.3}& 83.4 & \textbf{74.8} & \textbf{56.0}\\
    \bottomrule
    \end{tabular}
}
\end{center}
\caption{Video detection results on ImageNet-VID. }
\label{tab:detection}
\end{table}

\renewcommand \arraystretch{0.9}
\begin{table*}[t]
\small
\begin{center}
\begin{tabular}{lcccccc}
\toprule
& \textbf{Acc.$\uparrow$} &  \textbf{FLOPs$\downarrow$} &  \textbf{\#Param.$\downarrow$} &  \textbf{Input size$\downarrow$} & \textbf{Throughput$\uparrow$} & \textbf{Compute/IO$\uparrow$}\\  
\midrule
\itddd~\cite{hara2018can} & 68.0\% & 40G & 47.0M & 16$\times$3$\times$224\textsuperscript{2} & 63.1V/s (1.5$\times$) & 16.6k (2.4$\times$)\\
\itd~\cite{wang2017non} & 73.3\% & 33G & 29.3M & 32$\times$3$\times$224\textsuperscript{2} & 41.9V/s (1.0$\times$) & 6.85k (1$\times$)\\
\midrule
TSM & \textbf{74.1\%} & 33G & 24.3M & 8$\times$3$\times$224\textsuperscript{2} & 84.8V/s (\textbf{2.0$\times$}) & 27.4k (\textbf{4$\times$}) \\
\bottomrule 
\end{tabular}
\end{center}
\caption{Efficiency statistics of different models. Arrows show the better direction.}
\label{tab:compared_model}
\end{table*}

\begin{figure*}[t]
\centering
\begin{subfigure}[b]{0.3\textwidth}
        \centering
        \includegraphics[height=1.4in]{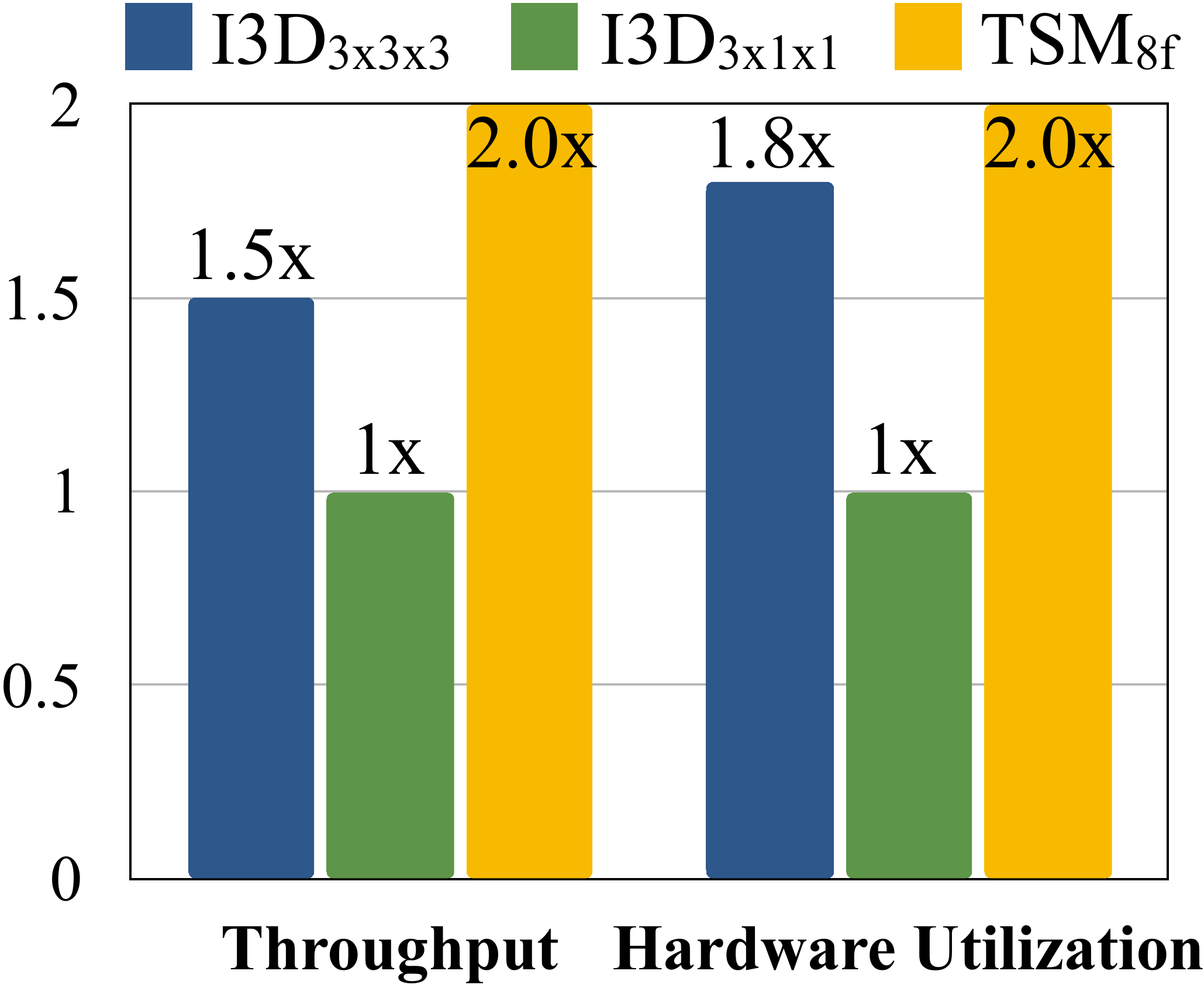}
        \caption{TSM has fewer FLOPs, better throughput and utilization.}
        \label{fig:hardware_efficiency}
    \end{subfigure}%
    ~ 
    \begin{subfigure}[b]{0.3\textwidth}
        \centering
        \includegraphics[height=1.3in]{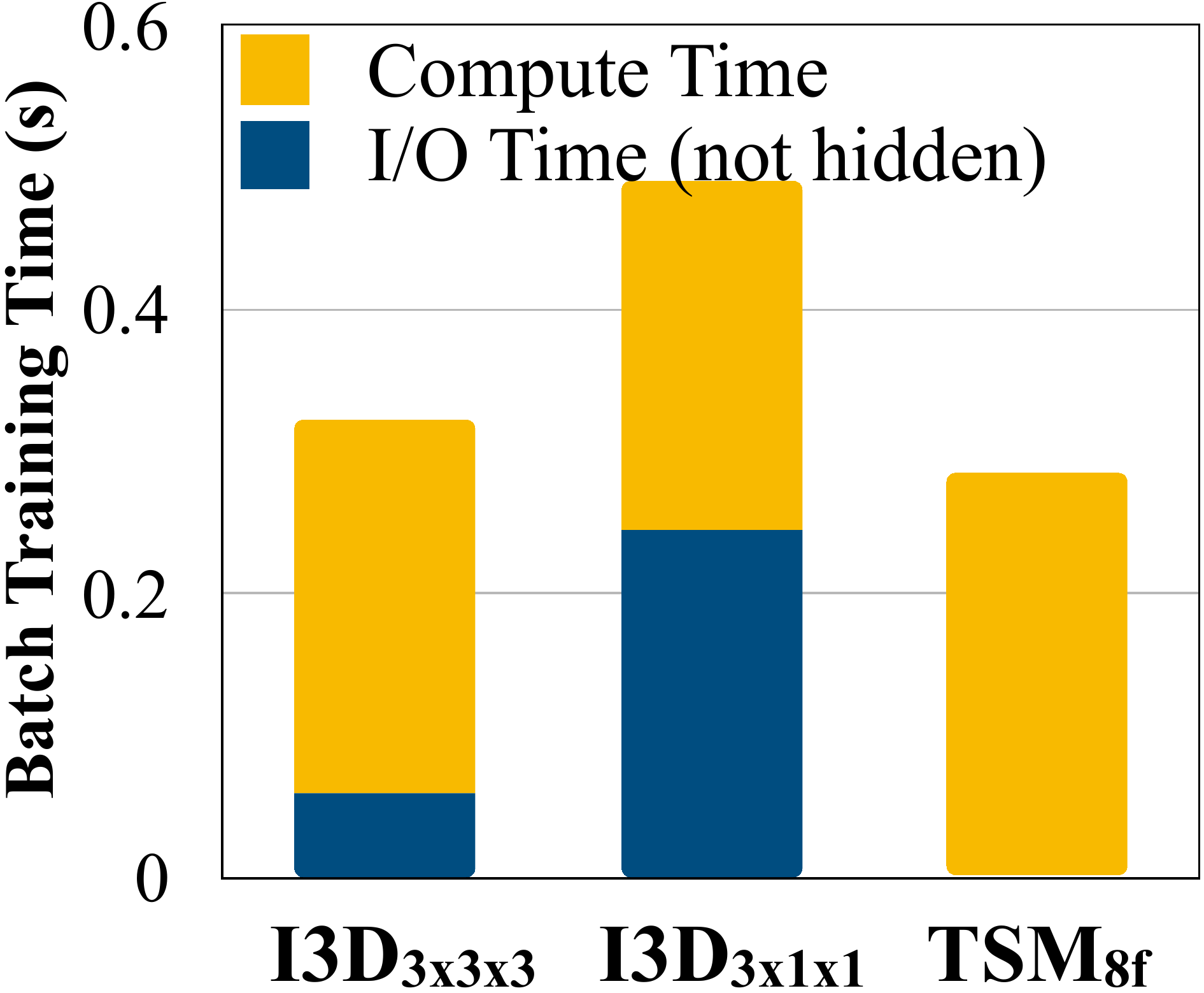}
        \caption{TSM is I/O light, decreasing the total batch time.}
        \label{fig:batch_timing}
    \end{subfigure}
     ~ 
    \begin{subfigure}[b]{0.3\textwidth}
        \centering
        \includegraphics[height=1.3in]{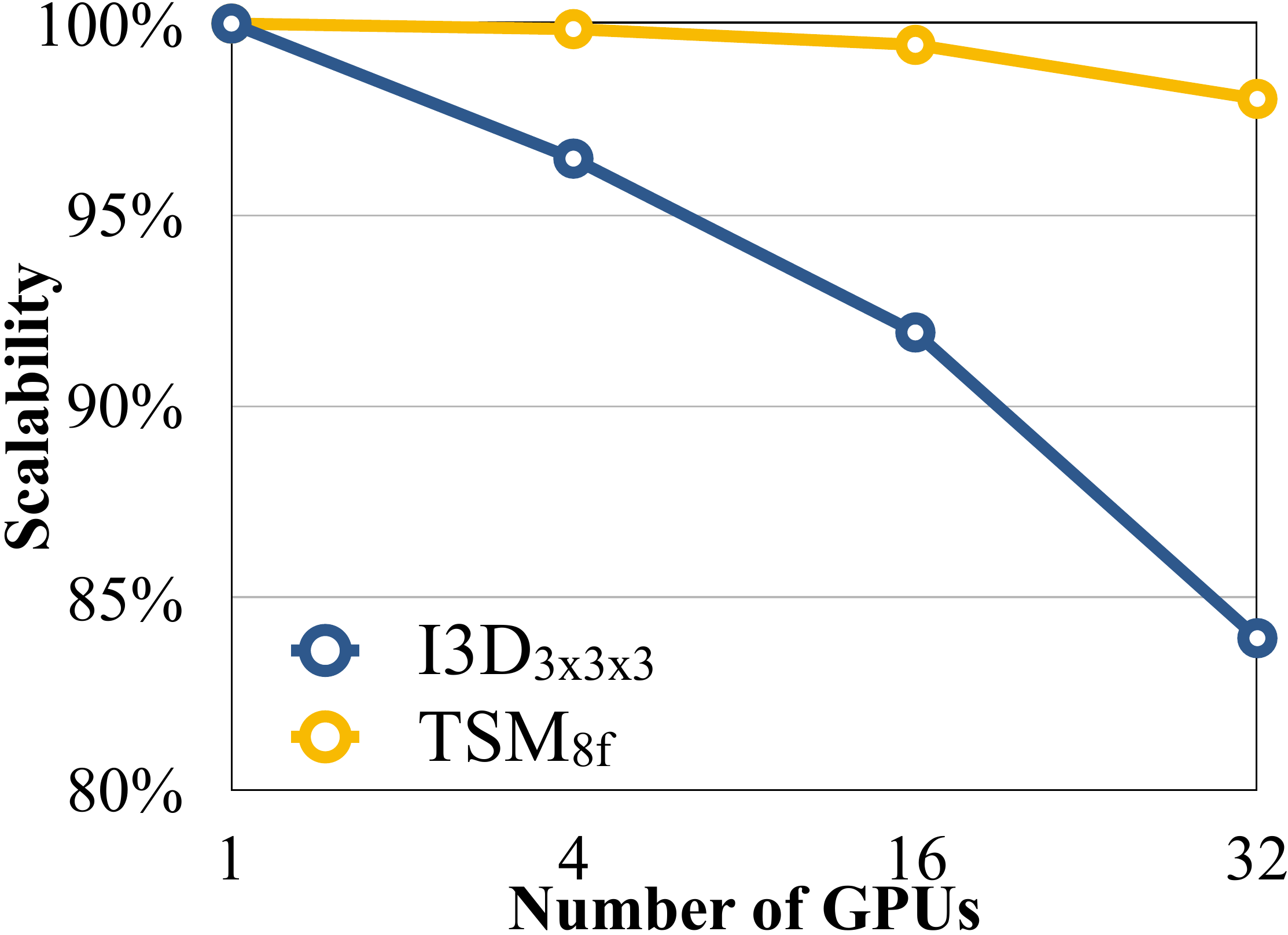}
        \caption{TSM has better scalability due to smaller model size.}
        \label{fig:model_scalability}
    \end{subfigure}
\caption{Analyzing how different design aspects influence the distributed training scalability of video recognition models: \textbf{(a) computation efficiency; (b) data loading efficiency; (c) networking efficiency}.}
\label{fig:guideline}
\end{figure*}

\begin{figure}[t]
\centering
    \begin{subfigure}[b]{0.2337\textwidth}
        \centering
        \includegraphics[width=\linewidth]{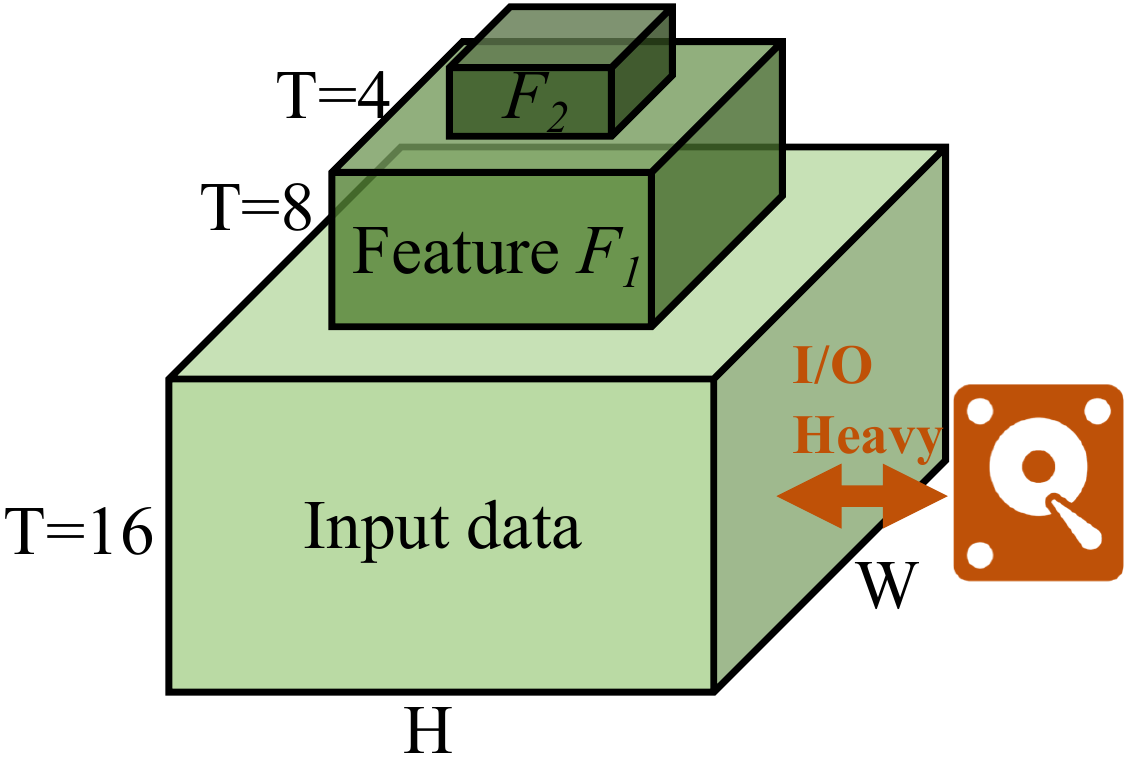}
        \caption{Pooled-up}
        \label{fig:pooledup}
    \end{subfigure}
    ~~
    \begin{subfigure}[b]{0.22\textwidth}
        \centering
        \includegraphics[width=\linewidth]{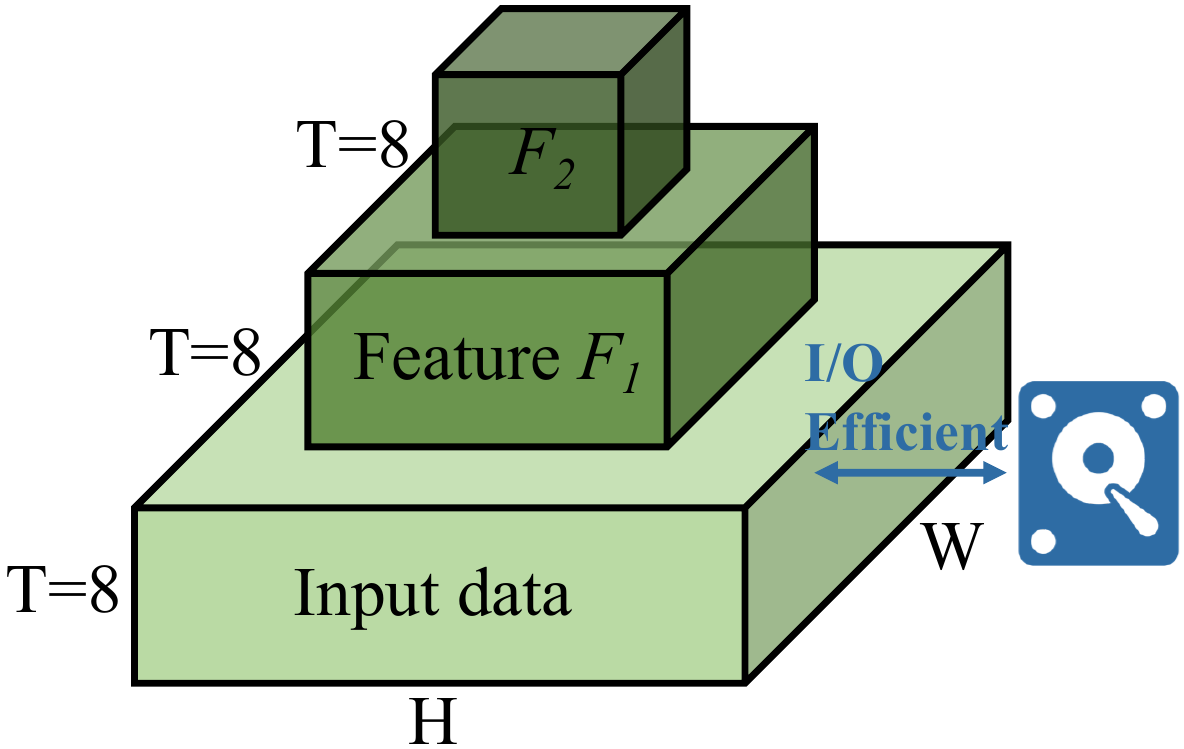}
        \caption{Straight-up (ours)}
        \label{fig:straightup}
    \end{subfigure}%
\caption{Two kinds of video backbone design. Straight-up backbone does not perform temporal pooling and is more data efficient. Pooled-up version requires many input frames and drains I/O.    }
\label{fig:test2}
\vspace{-15pt}
\end{figure}

\section{Scalability in Distributed Training}

In this section, we study how the design of TSM helps to improve the scalability in distributed training of video models.

\subsection{Factor of Video Network Design}

To study the distributed training scalability, we first discuss the factors that might affect the scalability of video network training~\cite{lin2019training}.

\subsubsection{Temporal modeling unit.} 
3D convolution is the most widely used operator for spatial-temporal modeling. However, it suffers from two problems: (1) large computation and large parameter size, which slows down training and communication; (2) low hardware efficiency compared to 2D convolution. Give the same amount of FLOPs, 3D kernels run 1.2 to 3 times slower than 2D on cuDNN~\cite{chetlur2014cudnn}. On the other hand, our TSM module is a highly efficient alternative.

\subsubsection{Backbone topology.}
Existing video networks usually sample \textbf{many} frames as input (32 frames~\cite{wang2017non} or 64 frames~\cite{carreira2017quo}), and perform temporal pooling later to progressively \textit{reduce} the temporal resolution (Figure~\ref{fig:pooledup}). Another way is to sample \textbf{fewer} frames (e.g. 8 frames~\cite{wang2016temporal}) as input while keeping the \textit{same} temporal resolution to keep the information (Figure~\ref{fig:straightup}). Although the overall computation of the two designs are similar, the former significantly increases the data loading traffic, making the system I/O heavy, which could be quite challenging in a distributed system considering the limited disk bandwidth.

\subsection{Design Guidelines to Video Model Architecture} \label{sec:guidelines}

To tackle the challenge in a distributed training systems, we propose three video model design guidelines: (1) To increase the \emph{computation efficiency}, use operators with lower FLOPs and higher hardware efficiency; (2) To reduce \emph{data loading traffic}, use a network topology with higher FLOPs/data ratio; (3) To reduce the \emph{networking traffic}, use operators with fewer parameters.

We show the advantage of the above three design guidelines by experimenting on three models in Table~\ref{tab:compared_model}. All the models use the ResNet-50 backbone to exclude the influence of spatial modeling. 
The model architectures are introduced as follows.

(1) The first model is an I3D model from~\cite{hara2018can}. The model takes 16 frames as input and inflate all the $3\times3$ convolutions to $3\times3\times3$. It performs temporal dimension pooling by four times to reduce the temporal resolution. We denote the model as \itddd. 

(2) The second model is an I3D model from~\cite{wang2017non}, taking 32 frames as input and inflating the first $1\times1$ convolution in every other ResBlock. It applies temporal dimension pooling by three times.
We denote this more computation and parameter efficient design as \itd. 

(3) The third model is built with TSM. The TSM operator is inserted into every ResBlock. The model takes 8 frames as input and performs no temporal pooling. We denote this model as TSM. 

\renewcommand \arraystretch{1.1}
\begin{table*}
\small
\begin{center}
\begin{tabular}{l|c|ccccccc|c}
\textbf{Block} & data & conv\textsubscript{1} & pool\textsubscript{1} & res\textsubscript{2} &  res\textsubscript{3} &  res\textsubscript{3} & res\textsubscript{4} &  res\textsubscript{5} & global pool \\
\hline
\textbf{\itddd} & 16 & 16 & 8 & 8 & - & 4 & 2 & 1 & 1 \\
\textbf{\itd} & 32 & 16 & 8 & 8 & 4 & 4 & 4 & 4 & 1 \\ \hline
\textbf{\tsm} & 8 & 8 & 8 & 8 & - & 8 & 8 & 8 & 1 \\
\end{tabular} 
\end{center}
\caption{The temporal resolution of output feature map for each block. 
TSM is a fully 2D structure, enjoying the best hardware efficiency. The last several stages of \itddd have fewer temporal resolution, making it more similar to 2D CNN, thus enjoying better hardware efficiency compared to \itd. }
\label{tab:network_structure}
\end{table*}

\subsubsection{Computation Efficiency.} Computation efficiency is the most direct factor that influence the training time. As shown in Table~\ref{tab:compared_model}, TSM\textsubscript{8f} has 1.2$\times$ fewer FLOPs compared to \itddd and roughly the same FLOPs compared to \itd. However, the actual inference throughput also depends on the hardware utilization. We measure the inference throughput (defined as videos per second) of the three models on a single NVIDIA Tesla P100 GPU using batch size 16. We also measured the hardware utilization, defined as achieved FLOPs/second over peak FLOPs/second. The inference throughput and the hardware efficiency comparison is shown in Figure~\ref{fig:hardware_efficiency}. We can find that the \emph{model is more hardware efficient if it has more 2D convolutions than 3D}: TSM is a fully 2D CNN, therefore it has the best hardware utilization (2.0$\times$); while the last several stage of \itddd (res\textsubscript{4}, res\textsubscript{5}) have few temporal resolution (as shown in Table~\ref{tab:network_structure}), it is more similar to 2D convolution and thus is 1.8$\times$ more hardware efficient than \itd  (1.0$\times$).

\subsubsection{Data Loading Efficiency.} Video datasets are usually very large. For a distributed system like the Summit supercomputer, the data is usually stored in High Performance Storage System (HPSS) shared across all the worker nodes. Such file systems usually have great sequential I/O performance but inferior random access performance. Therefore, large data traffic could easily become the system bottleneck. Previous popular I3D models~\cite{carreira2017quo, hara2018can} takes many frames per video (16 or 32) as input and perform down-sample over temporal dimension. We argue that such design is a waste of disk bandwidth: a TSM\textsubscript{8f} only takes 8 frames as input while achieving better accuracy. The intuition is that nearby frames are similar; loading too many similar frames is redundant. 
We empirically test the data loading bottleneck on Summit. To exclude the communication cost from the experiments, we perform timing on single-node training. We measure the total time of one-batch training and the time for data loading (that is not hidden by the computation). As shown in Figure~\ref{fig:batch_timing}, for \itd, it takes 32 frames as input. The data loading time cannot be hidden by the computation, therefore data I/O becomes the bottleneck. \itddd that takes 16 frame as input has less problem on data loading, while TSM\textsubscript{8f} can fully hide the data loading time with computation. We also compute the model FLOPs divided by the input data size as a measurement of data efficiency. The value is denoted as "Compute/IO" as in Table~\ref{tab:compared_model}. For scalable video recognition models, we want a model with larger Compute/IO ratio.

\subsubsection{Networking Efficiency.} In distributed training system, the communication time can be modelled as:
\begin{equation}
    \text{communication time} =  \text{latency} + \frac{\text{model size}}{\text{bandwidth}}
\end{equation}
The latency and bandwidth is determined by the network condition, which cannot be optimized through model design. However, we can reduce the model size to reduce the communication cost. Both I3D models inflate some of the 2D convolution kernels to 3D, which will increase the number of parameters by $k_T$. While TSM module does not introduce extra parameters. Therefore, it has the same model size as the 2D counterpart. For example, \itddd has 1.9$\times$ larger model size than TSM\textsubscript{8f}, which introduces almost two times of network communication during distributed training.
To test the influence of model size on scalability, we measure the scalability on a 8 node cluster. Each computer has 4 NVIDIA TESLA P100 GPUs. We define the scalability as the actual training speed divided by the ideal training speed (single machine training speed * number of nodes). The results are shown in Figure~\ref{fig:model_scalability}. Even with the high-speed connection, the scalability of \itddd quickly drops as the number of training nodes increase: the scalability is smaller than 85\% when applied to 8 nodes. While \tsm model still has over 98\% of scalability thanks to the smaller model size thus smaller networking traffic.

\subsection{Large-scale Distributed Training on Summit}

\begin{figure}[t]
\centering
\includegraphics[width=0.4\textwidth]{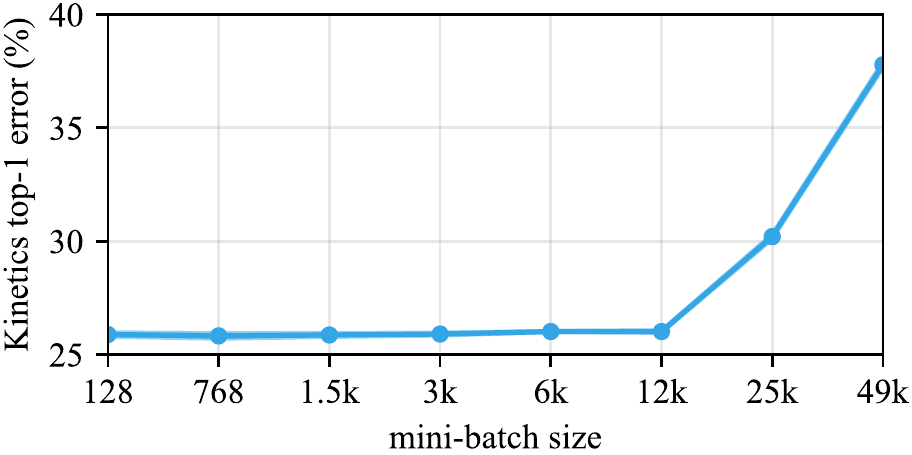}
\caption{Kinetics top-1 validation accuracy \vs mini-batch size. The performance of the model does not degrade when we scale up the mini-batch size to $12k$. The mean and standard deviation (the scale of the STD is hardly visible) are shown in the figure.}
\label{fig:err_vs_bs}
\end{figure}
\renewcommand \arraystretch{0.95}
\begin{table}[t]
\setlength{\tabcolsep}{2pt}
\small
\begin{center}
\begin{tabular}{lllllll}
\toprule
\textbf{\#Node} &  \textbf{\#GPU} &  \textbf{Batch} & \textbf{\#Frames} &  \textbf{Accuracy} &  \textbf{Time} & \textbf{\emph{Note}}\\
\midrule
1 & 6 & 96 & 768 & 74.12$\pm$0.11 & 49h 55m & \emph{Baseline}\\
\midrule
8 & 48 & 384 & 3,072 & 74.12$\pm$0.08 & 7h 7m & \multirow{6}{*}{\emph{\shortstack{same level \\ of \\ accuracy}}}\\
16 & 96 & 768 & 6,144 & 74.18$\pm$0.14 & 3h 38m &  \\
32 & 192 & 1,536 & 12,288 &  74.14$\pm$0.10 &  1h 50m  \\
64 & 384 & 3,072 & 24,576 & 74.10$\pm$0.08 & 55m 56s\\
128 & 768 & 6,144 & 49,152 & 73.99$\pm$0.04 & 28m 14s\\
256 & 1536 & 12,288 & 98,304 & 73.99$\pm$0.07 & 14m 13s \\
\midrule
384 & 2304 & 18,432 & 147,456 & 72.52$\pm$0.07 & 10m 9s & \multirow{3}{*}{\emph{\shortstack{lose \\ accuracy}}}\\
512* & 3072 & 24,576 & 196,608 & 69.80$\pm$0.13 & - \\
1024* & 6144 & 49,152 & 393,216 & 62.22$\pm$0.17 & -\\
\bottomrule 
\end{tabular}
\end{center}
\caption{Detailed statistics of different mini-batch size (* indicates simulated performance). }
\label{tab:err_vs_bs}
\end{table}

\begin{figure}[t]
\centering
\includegraphics[width=0.48\textwidth]{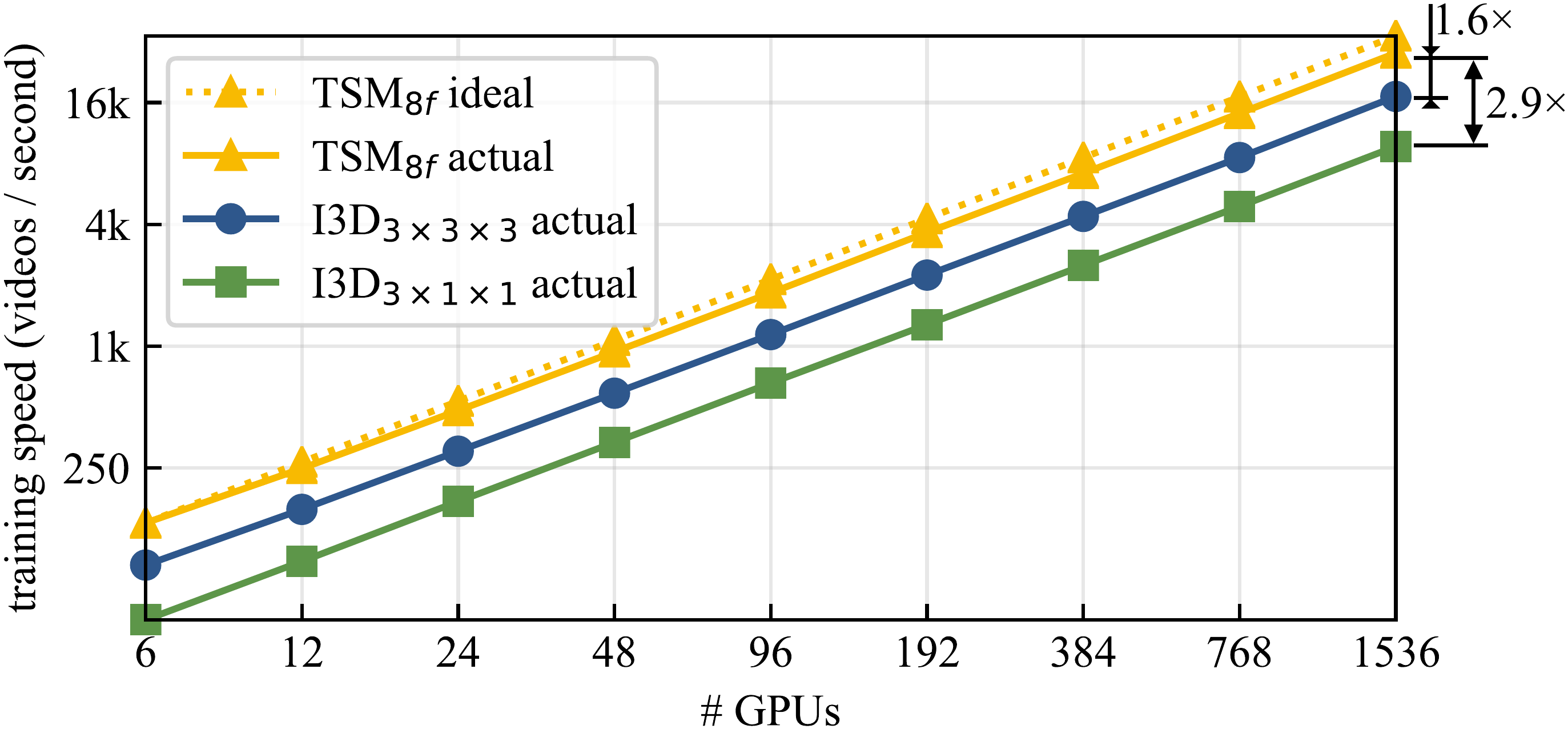}
\caption{The training speed and scalability of distributed synchronous SGD training. \tsm achieves a good scalability ($>$80\%) even when using 1536 GPUs. 
\tsm can achieve 1.6$\times$ higher training speed compared to \itddd and 2.9$\times$ compared to \itd, showing the effectiveness of the proposed design guidelines.}
\label{fig:scalability}
\vspace{-5pt}
\end{figure}

\begin{figure*}
\centering
\begin{subfigure}[b]{0.3\textwidth}
	\includegraphics[width=\textwidth]{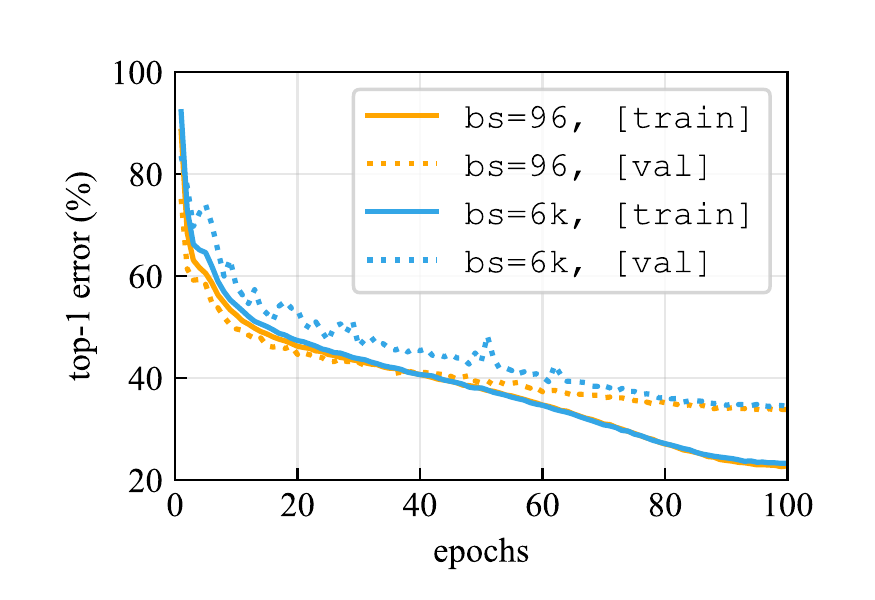}
	\caption{Mini-batch size $6k$.}
	\label{fig:curve6k}
\end{subfigure}
\begin{subfigure}[b]{0.3\textwidth}
	\includegraphics[width=\textwidth]{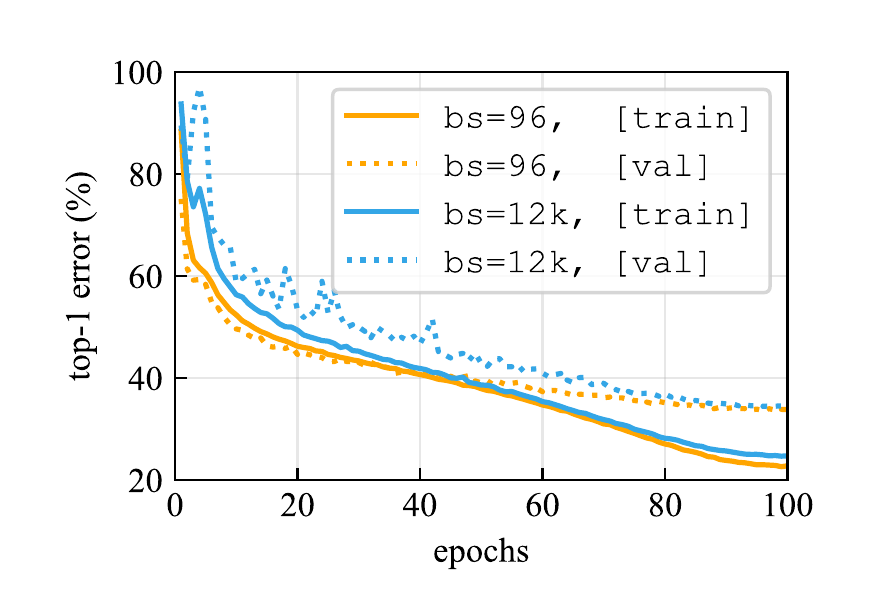}
	\caption{Mini-batch size $12k$.}
	\label{fig:curve12k}
\end{subfigure}
\begin{subfigure}[b]{0.3\textwidth}
	\includegraphics[width=\textwidth]{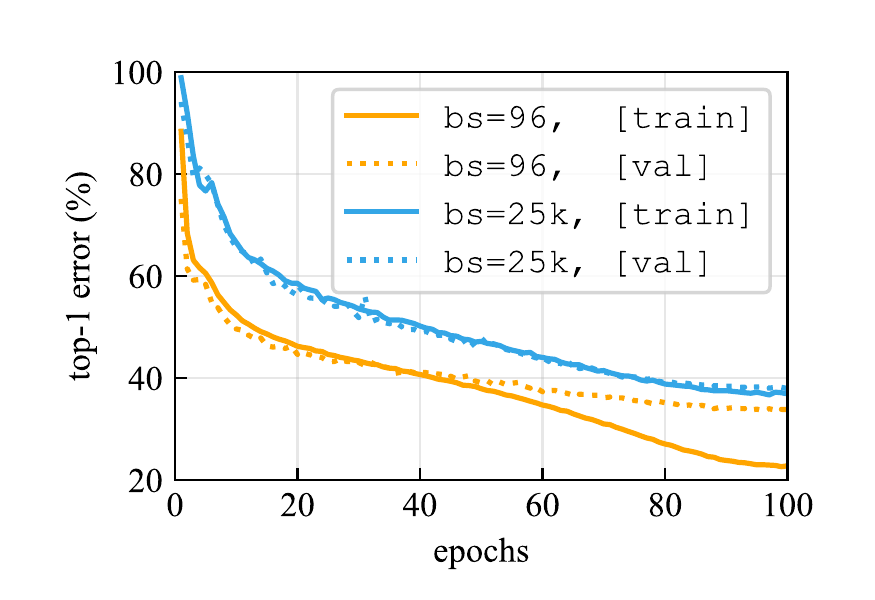}
	\caption{Mini-batch size $25k$ (degrade).}
	\label{fig:curve25k}
\end{subfigure}
\vspace{-5pt}
\caption{The learning curve for baseline training and large-batch distributed training (batch size 6144, 12228, 24576). The performance does not degrade for batch size 6144 and 12228, while degrades for a batch size of 24576.}
\label{fig:training_curve}
\vspace{10pt}
\includegraphics[width=0.85\textwidth]{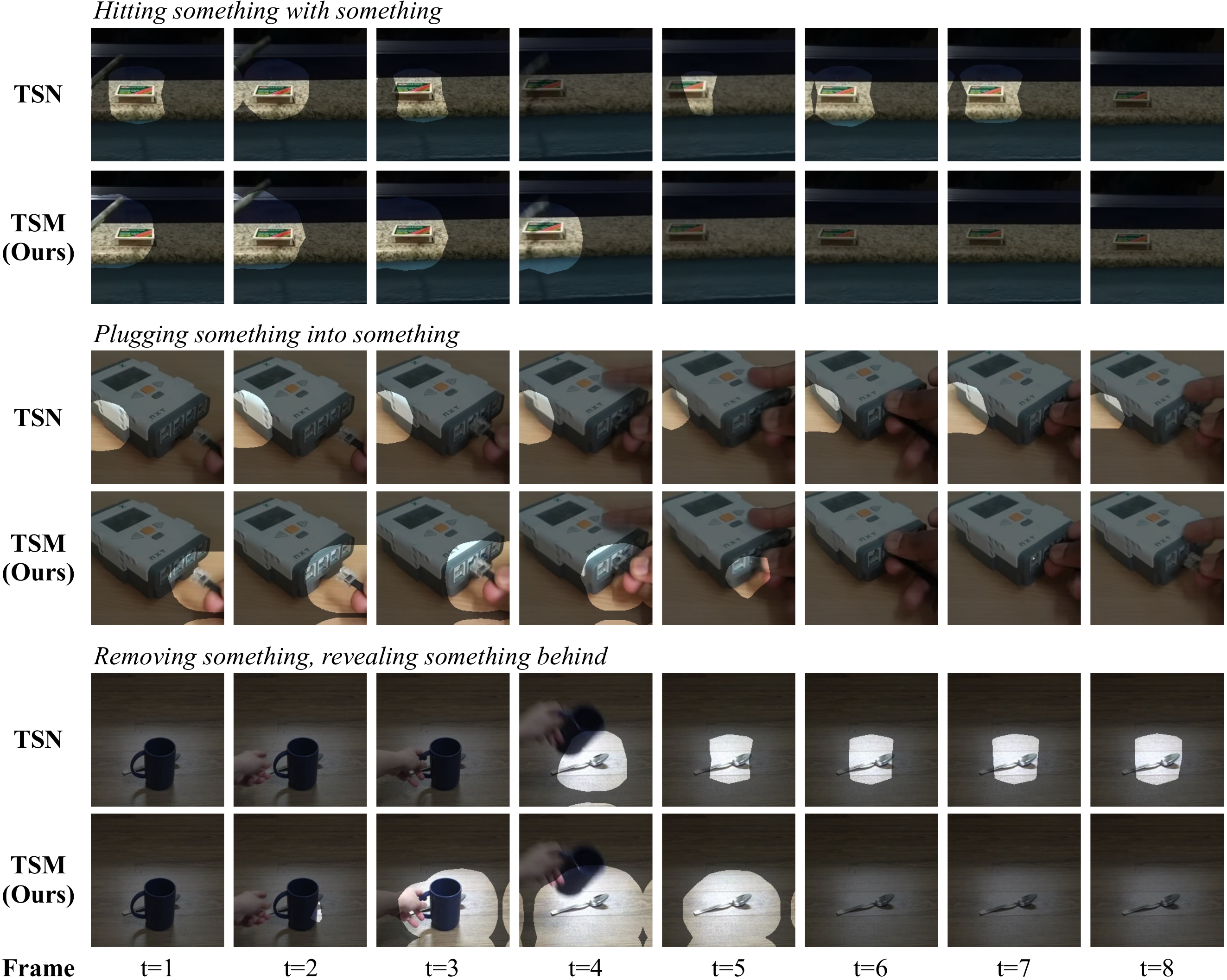}
\caption{Spatial-temporal action detector emerges in TSM video classification models, while single-frame baseline (TSN) cannot localize the action. \emph{Italic} title indicates the action category. In the first example, our TSM model precisely localize the ``hitting'' action, while TSN can only highlight the object. In the second example, TSM localizes the ``plugging'' action but not the other hand motion. Finally, TSM accurately locates the temporal region where the ``removing'' action happens. }
\label{fig:dissection}
\end{figure*}

We scale up the training of video recognition model on Summit supercomputer. With the help of above hardware-aware model design techniques, we can scale up the training to 1536 GPUs, finishing the training of Kinetics in 15 minutes.

\subsubsection{Setups}

Summit~\cite{vazhkudai2018design} or OLCF-4 is a supercomputer at Oak Ridge National Laboratory, which as of September 2019 is the fastest supercomputer in the world. It consists of approximately 4,600 compute nodes, each with two IBM POWER9 processors and six NVIDIA Volta V100 accelerators. The POWER9 processor is connected via dual NVLINK bricks, each capable of a 25GB/s transfer rate in each direction. Nodes contain 512 GB of DDR4 memory for use by the POWER9 processors and 96 GB of High Bandwidth Memory (HBM2) for use by the accelerators~\footnote{\url{https://www.olcf.ornl.gov/for-users/system-user-guides/summit/summit-user-guide}}.

We used PyTorch and Horovod~\cite{sergeev2018horovod} for distributed training. The framework uses ring-allreduce algorithm to perform synchronized SGD. The training is accelerated by CUDA and cuDNN. 
We used NVIDIA Collective Communication Library (NCCL)~\footnote{\url{https://developer.nvidia.com/nccl}} for most of the communication. 

For training on Kinetics, we used the same hyper-parameter at the same batch size, applying a linear scaling rule~\cite{goyal2017accurate}.

\subsubsection{Experiments}

\noindent\textbf{Baseline.}
For the baseline, we trained a ResNet-50 \tsm model on a single Summit node with 6 GPUs, each GPU contains 16 video clips, resulting in a total batch size of $kn=96$. We evaluate the performance of last 5 checkpoints, it achieves a top-1 accuracy of $74.12\pm 0.11\%$.

\noindent\textbf{Performance \vs Batch Size.}
We first compare the training error \vs the batch size. As shown in~\cite{goyal2017accurate}, the accuracy will not degrade when the batch size is relatively small. Therefore, our experiments start from 8 computing nodes (48 GPUs, 384 video clips, 3072 frames) to 1024 computing nodes (6144 GPUs, 49152 video clips, 393216 frames) . Note that each sample in a video recognition model is a video clip consisting of several frames/images (in our case, 8). Therefore, the actual number of images used in one batch could be much larger than ImageNet training (\eg, $98k$ \vs $8k$~\cite{goyal2017accurate}).

We first plot the error \vs batch size trade-off in Figure~\ref{fig:err_vs_bs}. The error does not increase when we scale the number of computing nodes up to 256 (1536 GPU), where the batch size is 12288, the total frame number is 98304. The detailed statistics are shown in Table~\ref{tab:err_vs_bs}. The scalability of TSM model is very close to the ideal case. Note that due to quota limitation, the largest physical nodes we used is 384 with 2304 GPUs. For 512 and 1024 nodes, we used gradient accumulation to simulate the training process (denoted by *).

We also provide the training and testing convergence curves using 768, 1536 and 3072 GPUs in Figure~\ref{fig:training_curve}. For 768 GPUs and 1536 GPUs, although the convergence of large-batch distributed training is slower than single-machine training baseline, the final converged accuracy is similar, so that the model does not lose accuracy. For 3072 GPUs, the accuracy degrades for both training and testing.

\noindent\textbf{Scalabilty.}
We test the scalability of distributed training on Summit. According to the results from last section, we can keep the accuracy all the way to 256 computing nodes. Therefore, we sweep the number of computing nodes from 1 to 256 to measure the scalability. We keep a batch size of 8 for each GPU and each node has 6 GPUs. So the batch sizes change from 48 to 18,432. Each video clips contains 8 frames in our model, resulting a total number of frames from 384 to 147,456.
We measure the training speed (videos/second) to get the actual speed-up. We calculate the ideal training speed using the single node training speed multiplied by number of nodes. The comparison of different models is provided in Figure~\ref{fig:scalability}. The actual training speed is just marginally below the ideal scaling, achieving $>80\%$ scaling efficiency. 
We also provide the detailed overall training time in Table~\ref{tab:err_vs_bs}.
With 1536 GPUs, we can finish the Kinetics training with TSM within 14 minutes and 13 seconds, achieving a top-1 accuracy of $74.0\%$. The overall training speed of \tsm is 1.6$\times$ larger than \itddd and 2.9$\times$ larger than \itd, showing the advantage of hardware-aware model design.

\section{Video Network Dissection}

In this section, we dissect the trained TSM model to understand how it learns temporal information compared to 2D networks.

To investigate what the action recognition network is learning, we adopt a similar method as in~\cite{zhou2016learning} to get the Class Activation Mapping (CAM), which shows the salience of each class over the input image. Take ResNet backbone as an example, the original output for the video network is:
\begin{equation}
    \text{logit} = \text{fc}(\text{pool}(x_{conv}))
\end{equation}

where $x_{conv}$ is the output activation of the last convolutional layer (\eg, has shape $1\times2048\times8\times7\times7$), pool is the global average pooling over both spatial and temporal dimension (reduce the shape to $1\times2048$), and fc is the last fully connected layer for classification. To get CAM, we remove the global average pooling layer, and change the fc layer to a $1\times1\times1$ convolution using the same weights, which results in a output tensor of shape $1\times\#\text{class}\times8\times7\times7$. We use the CAM map of the predicted category (highest probability) as the attention of the network.

For visualization, we used a similar method as in~\cite{bau2017network}. Specifically, we first resize the spatial resolution of CAM feature map to the size of the input video clip ($1\times\#\text{class}\times8\times224\times224$) with bilinear interpolation and use a threshold to divide the attention foreground and background. We set the threshold to preserve 20\% of the pixels over the validation set.

We perform experiments on Something-Something V2~\cite{goyal2017something} dataset. And some results are shown in Figure~\ref{fig:dissection}. We compare the attention distribution between our TSM model and 2D baseline TSN. The background of the category-aware attention map is darkened. We find that spatial-temporal action detector emerges in TSM video network, even though we only provide classification label during the training. TSM models can accurately localize the ``action'', instead of the ``object''. For example, in the first video clip labeled as ``Hitting something with something'', TSM model only highlights the region where a pen is hitting the card box, \ie, when and where the action is happening. However, for the 2D baseline, since it does not have the temporal information, it only highlights the object box. The same situation happens for the following two video clips. Note that in the third clip (``Removing something, revealing something behind''), the 5-th frame and the 6-th frame look exactly the same, while with the help of the temporal modeling, TSM model can tell that the 5-th frame is part of the action while the 6-th frame not.

\section{Conclusion}
We propose Temporal Shift Module for hardware-efficient video recognition. It can be inserted into 2D CNN backbone to enable joint spatial-temporal modeling at no additional cost. The module shifts part of the channels along temporal dimension to exchange information with neighboring frames. Our framework is both efficient and accurate, enabling low-latency video recognition on edge devices. It has better scalability than 3D networks, enabling large-scale training on video recognition. We also show that spatial-temporal action detector emerges in TSM network. 

{\small
\bibliographystyle{plain}
\bibliography{egbib}
}



\ifCLASSOPTIONcompsoc
  \section*{Acknowledgments}
\else
  \section*{Acknowledgment}
\fi

We thank MIT Quest for Intelligence, MIT-IBM Watson AI Lab, MIT-SenseTime Alliance, Samsung, SONY, AWS, Google for supporting this research. We thank Oak Ridge National Lab for Summit supercomputer. We thank John Cohn for the support for our work.

\ifCLASSOPTIONcaptionsoff
  \newpage
\fi



%

\newpage
\begin{IEEEbiography}[{\includegraphics[width=1in,height=1.25in,clip,keepaspectratio]{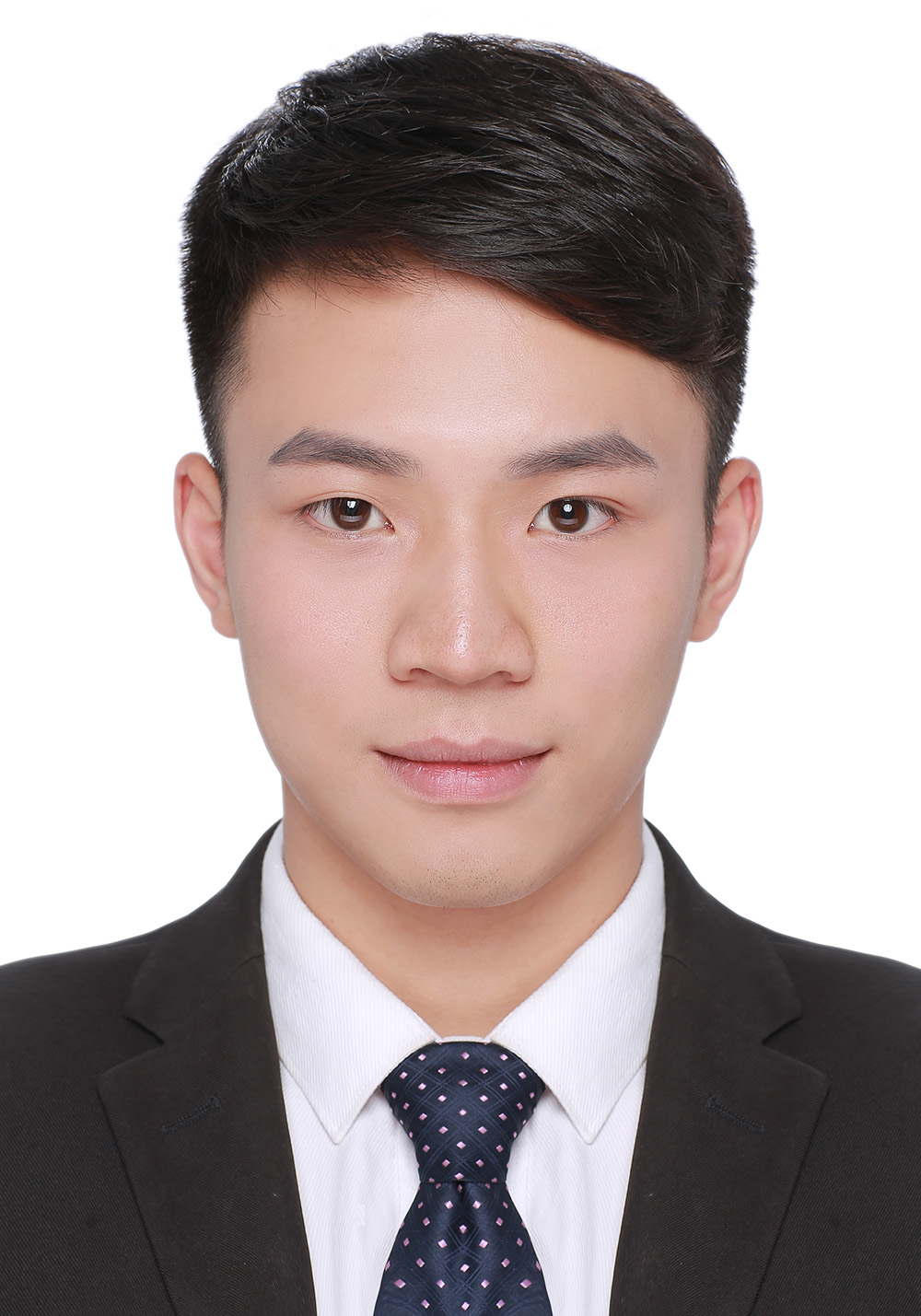}}]{Ji Lin}
 is currently a second-year Ph.D. student at MIT EECS. Previously, he graduated from Department of Electronic Engineering, Tsinghua University. His research interests lie in efficient and hardware-friendly machine learning and its applications.
\end{IEEEbiography}


\begin{IEEEbiography}[{\includegraphics[width=1in,height=1.25in,clip,keepaspectratio]{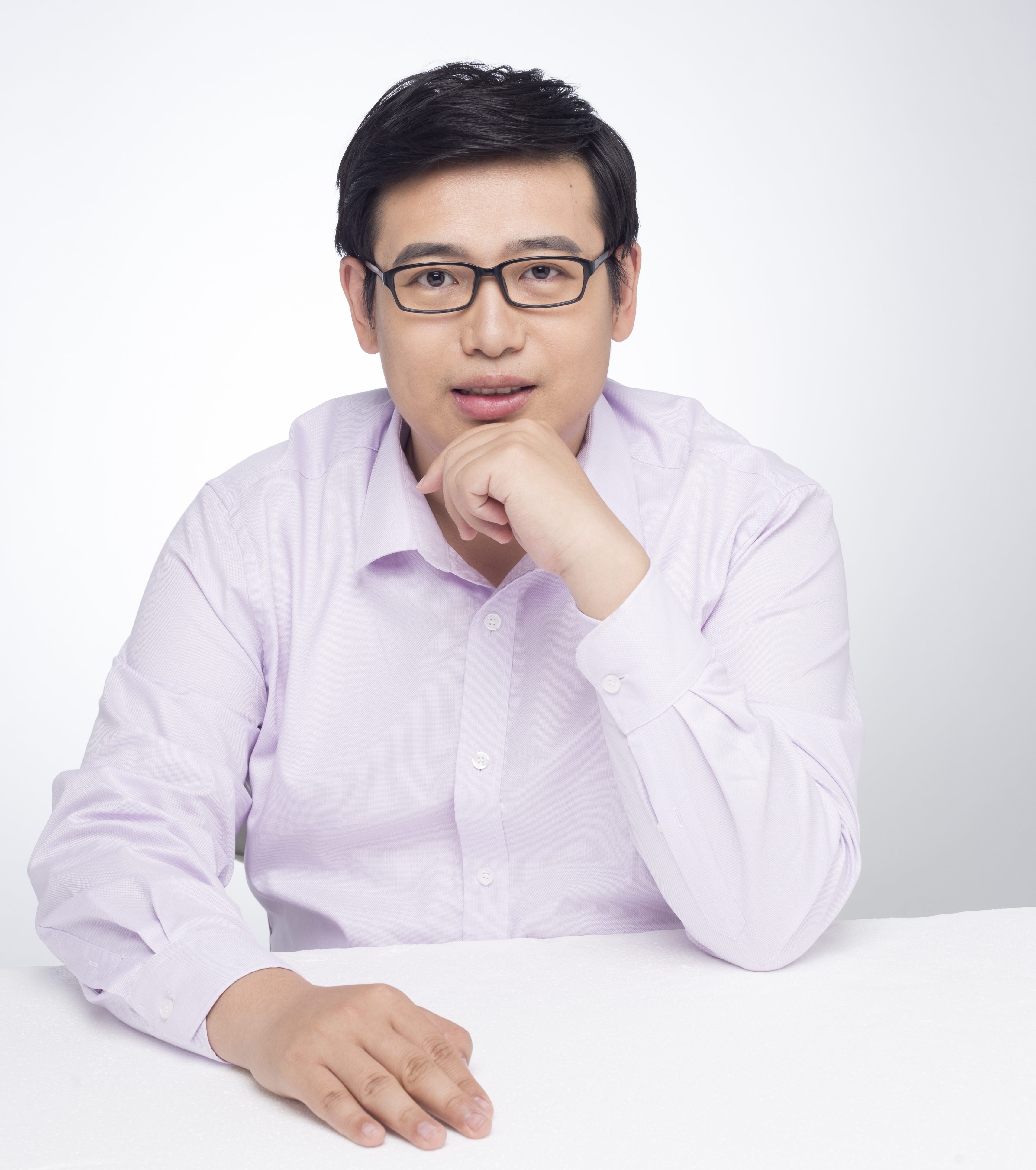}}]{Chuang Gan}
is a research staff member at MIT-IBM Watson AI Lab. He is also an affiliated researcher at MIT EECS. His research interests focus on computer vision and machine learning.
\end{IEEEbiography}


\begin{IEEEbiography}[{\includegraphics[width=1in,height=1.25in,clip,keepaspectratio]{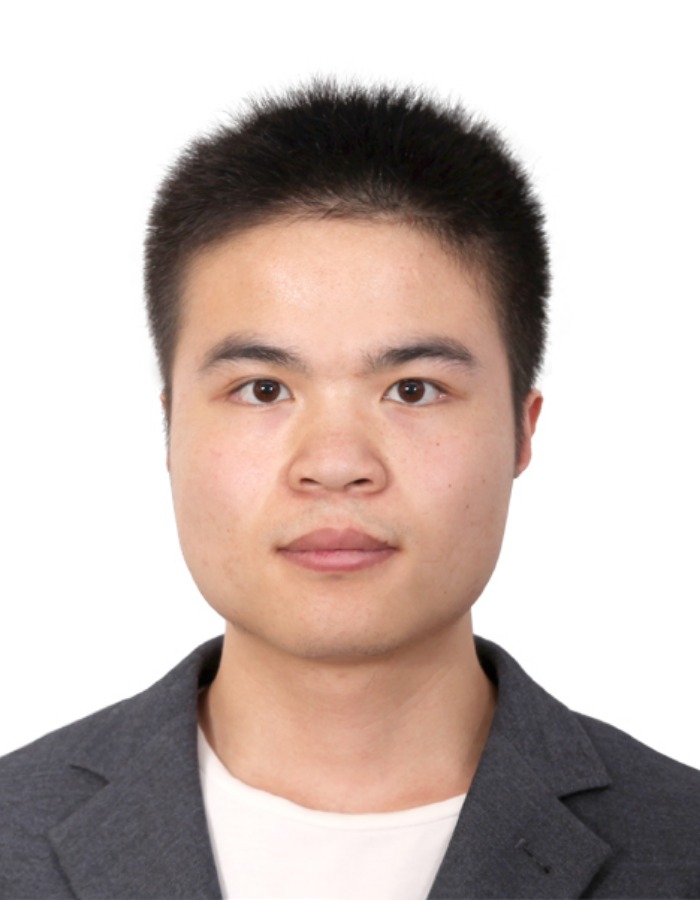}}]{Kuan Wang}
is a fourth-year undergraduate degree at Tsinghua University, and a visiting student at MIT, advised by Dr. Song Han. His current research interests lie on the intersection of computer vision, deep learning and efficient hardware architecture. He is a student member of the IEEE.
\end{IEEEbiography}


\begin{IEEEbiography}[{\includegraphics[width=1in,height=1.25in,clip,keepaspectratio]{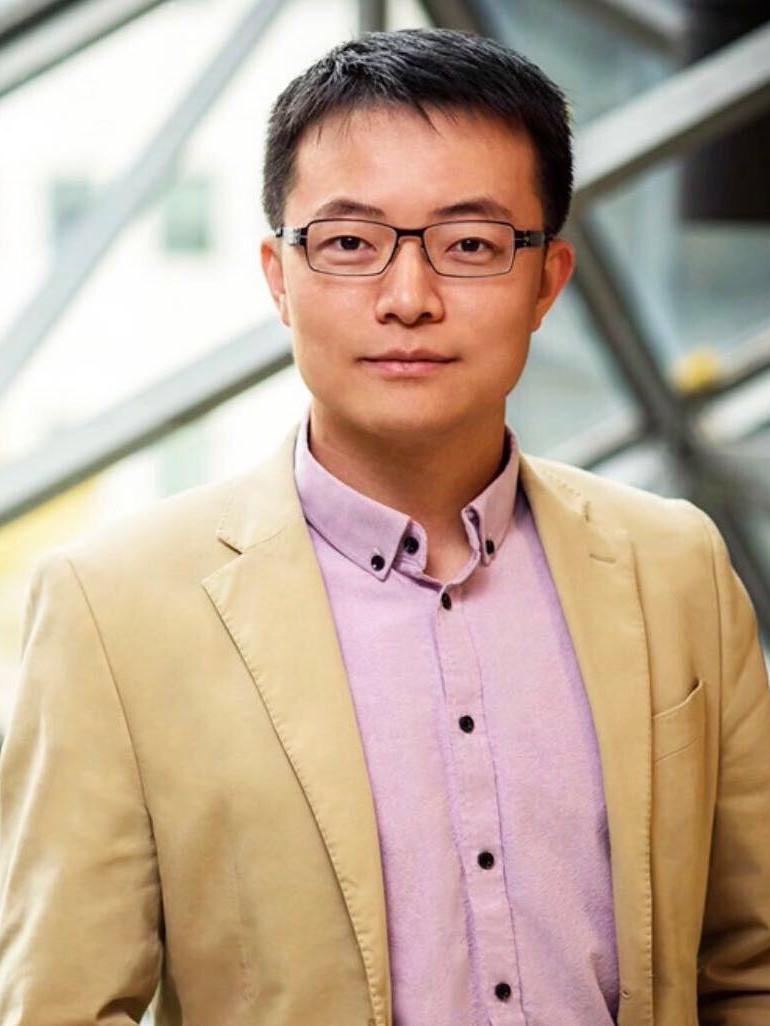}}]{Song Han}
is an assistant professor at MIT EECS Department. Dr. Han received the Ph.D. degree in Electrical Engineering from Stanford University and B.S. degree in Electrical Engineering from Tsinghua University. Dr. Han's research focuses on efficient deep learning computing at the intersection between machine learning and computer architecture. He proposed ``Deep Compression'' and the ``Efficient Inference Engine'' that impacted the industry. He is a recipient of NSF CAREER Award, MIT Technology Review Innovators Under 35, best paper award at the ICLR’16 and FPGA’17, Facebook Faculty Award, SONY Faculty Award, AWS Machine Learning Research Award. Contact: songhan@mit.edu. 
\end{IEEEbiography}




\end{document}


\title{Supplementary Material for \\ TSM: Temporal Shift Module for Efficient Video Understanding} 

\author{Ji Lin\\
MIT\\
{\tt\small jilin@mit.edu}
\and
Chuang Gan\\
MIT-IBM Watson AI Lab\\
{\tt\small ganchuang@csail.mit.edu}
\and
Song Han\\
MIT\\
{\tt\small songhan@mit.edu}
}

\maketitle

\section{Uni-directional TSM for Online Video Detection}

In this section, we show more details about the online video object detection with uni-directional TSM.

Object detection suffers from poor object appearance due to motion blur, occlusion, defocus, \etc. Video based object detection gives chances to correct such errors by aggregating and inferring temporal information. 

Existing methods on video object detection ~\cite{zhu2017flow} fuses information along temporal dimension after the feature is extracted by the backbone. 
Here we show that we can enable temporal fusion in online video object detection by injecting our uni-directional TSM into the backbone. We show that we can significantly improve the performance of video detection by simply modifying the backbone with online TSM, without changing the detection module design or using optical flow features.

We conducted experiments with R-FCN~\cite{dai16rfcn} detector on ImageNet-VID~\cite{russakovsky2015imagenet} dataset. Following the setting in~\cite{zhu2017flow}, we used ResNet-101~\cite{he2016deep} as the backbone for R-FCN detector. For TSM experiments, we inserted uni-directional TSM to the backbone, while keeping other settings the same. We used the official training code of~\cite{zhu2017flow} to conduct the experiments, and the results are shown in Table~\ref{tab:detection}. 
Compared to 2D baseline R-FCN~\cite{dai16rfcn}, our online TSM model significantly improves the performance, especially on the fast moving objects, where TSM increases mAP by $4.6\%$. 
FGFA~\cite{zhu2017flow} is a strong baseline that uses optical flow to aggregate the temporal information from 21 frames (past 10 frames and future 10 frames) for offline video detection. Compared to FGFA, TSM can achieve similar or higher performance while enabling online recognition (using information from only past frames) at much smaller latency per frame.
The latency overhead of TSM module itself is less than 1ms per frame, making it a practical tool for real deployment.

\begin{figure*}[t]
\centering
\includegraphics[width=0.95\textwidth]{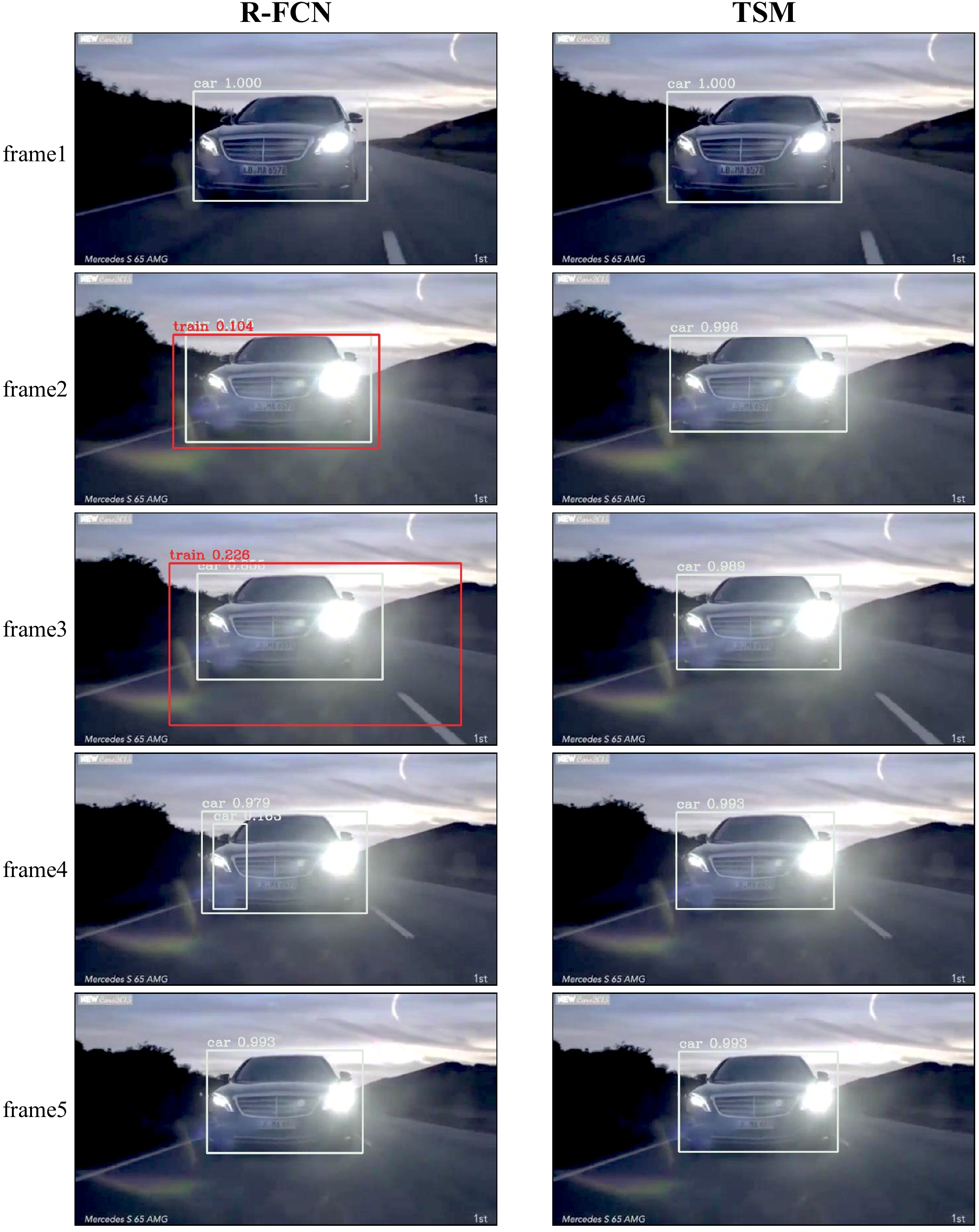}
\caption{Comparing the result of R-FCN baseline and TSM model. 2D baseline R-FCN generates false positive due to the glare of car headlight on frame 2/3/4, while TSM does not have such issue by considering the temporal information. }
\label{fig:det_car}
\end{figure*}

\begin{figure*}[t]
\centering
\includegraphics[width=0.95\textwidth]{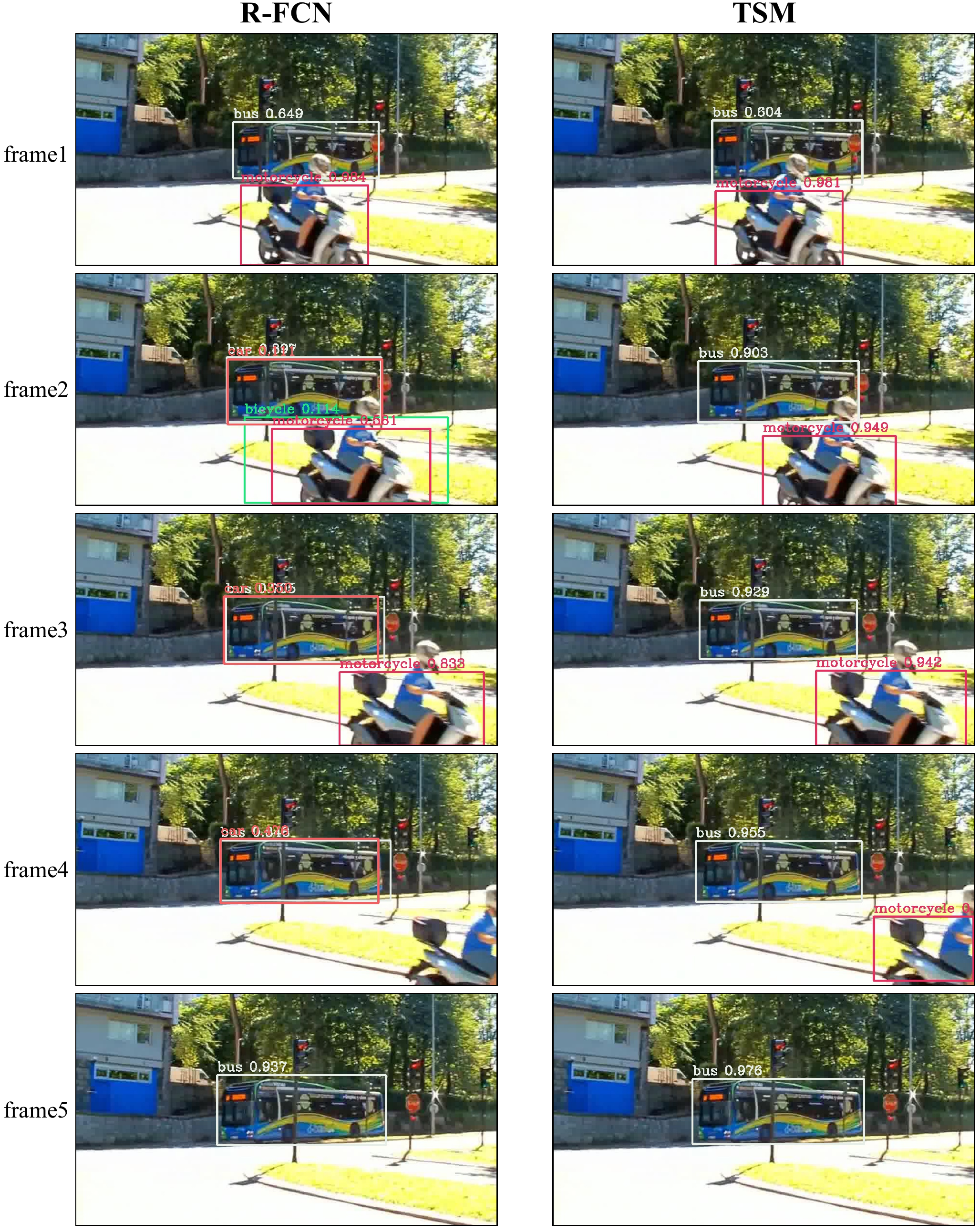}
\caption{Comparing the result of R-FCN baseline and TSM model. 2D baseline R-FCN generates false positive surrounding the bus due to occlusion by the traffic sign on frame 2/3/4. Also, it fails to detect motorcycle on frame 4 due to occlusion. TSM model addresses such issues with the help of temporal information. }
\label{fig:det_bus}
\end{figure*}

We visualize two video clips in Figure~\ref{fig:det_car} and~\ref{fig:det_bus}. In Figure~\ref{fig:det_car}, 2D baseline R-FCN generates false positive due to the glare of car headlight on frame 2/3/4, while TSM suppresses false positive. In Figure~\ref{fig:det_bus}, R-FCN generates false positive surrounding the bus due to occlusion by the traffic sign on frame 2/3/4. Also, it fails to detect motorcycle on frame 4 due to occlusion. TSM model addresses such issues with the help of temporal information. 

\renewcommand \arraystretch{0.9}
\begin{table}[t]
\setlength{\tabcolsep}{2.5pt}
\begin{center}
\footnotesize{
    \begin{tabular}{cccccccc}
    \toprule
    \multirow{2}{*}{\textbf{Model}} & \multirow{2}{*}{\textbf{Online}} & \multirow{2}{*}{\textbf{\shortstack{Need\\Flow}}} & \multirow{2}{*}{\textbf{Latency}} &
    \multicolumn{4}{c}{\textbf{mAP}} \\ 
    \cmidrule(lr){5-8}
    & & & & \textbf{Overall} & \textbf{Slow} & \textbf{Medium} & \textbf{Fast} \\
    \midrule
    R-FCN~\cite{dai16rfcn} & \checkmark & & 1$\times$ & 74.7 & 83.6 & 72.5 & 51.4\\
    FGFA~\cite{zhu2017flow} & & \checkmark & 2.5$\times$  & 75.9 & \textbf{84.0} & 74.4 & 55.6  \\
    \midrule
    Online TSM & \checkmark &  & 1$\times$ & \textbf{76.3}& 83.4 & \textbf{74.8} & \textbf{56.0}\\
    \bottomrule
    \end{tabular}
}
\end{center}
\caption{Video detection results on ImageNet-VID. }
\label{tab:detection}
\end{table}

\section{Video Demo}

We provide more video demos of our TSM model in the following project page: \url{https://hanlab.mit.edu/projects/tsm/}.

{\small
\bibliographystyle{ieee}
\bibliography{egbib}
}

\newpage

\newpage

